\documentclass{article}

\usepackage{arxiv}

\usepackage[utf8]{inputenc} 
\usepackage[T1]{fontenc}    
\usepackage{hyperref}       
\usepackage{url}            
\usepackage{booktabs}       
\usepackage{amsfonts}       
\usepackage{nicefrac}       
\usepackage{microtype}      
\usepackage{lipsum}
\usepackage{graphicx}
\graphicspath{ {./images/} }
\usepackage{tabularx}
\usepackage{float} 
\usepackage[utf8]{inputenc}
\usepackage[T1]{fontenc}
\usepackage{etoolbox}
\usepackage{textgreek}

\usepackage{textcomp} 
\usepackage{listings}
\usepackage{geometry}
\usepackage{upquote} 

\usepackage{fancyvrb} 
\usepackage{fvextra}  

\usepackage{xcolor}
\usepackage[many]{tcolorbox} 

\newcounter{suppfigure}
\renewcommand{\thesuppfigure}{S\arabic{suppfigure}}

\newenvironment{suppfigureenv}{
  \refstepcounter{suppfigure}%
  \begin{figure}[H]
}{
  \end{figure}
}

\newcommand{\suppcaptionbox}[2]{%
  \refstepcounter{suppfigure}%
  \textbf{Supplementary Figure \thesuppfigure.} #2%
  \label{#1}%
}

\title{Robin: a multi-agent system for automating scientific discovery}

\author{
  Ali Essam Ghareeb\textsuperscript{*, 1}
  \And
  Benjamin Chang\textsuperscript{*,1,2}
  \And
  Ludovico Mitchener\textsuperscript{1}
  \And
  Angela Yiu\textsuperscript{1}
  \And
  Caralyn J. Szostkiewicz\textsuperscript{1}
  \And
  Jon M. Laurent\textsuperscript{1}
  \And
  Muhammed T. Razzak 
  \And
  Andrew D. White\textsuperscript{1, $\dagger$}
  \And
  Michaela M. Hinks\textsuperscript{1, $\ddagger$}
  \And
  Samuel G. Rodriques\textsuperscript{1, $\dagger$, $\ddagger$}
}

\begin{document}
\maketitle

\begingroup 
\makeatletter
\renewcommand\@makefnmark{} 
\let\oldthefootnote\thefootnote
\renewcommand\thefootnote{} 
\footnotetext{\textsuperscript{*}Equal contribution.}
\footnotetext{\textsuperscript{1}FutureHouse, San Francisco, USA.}
\footnotetext{\textsuperscript{2}University of Oxford, Oxford, UK.}
\footnotetext{\textsuperscript{$\dagger$}These authors jointly supervise work at FutureHouse.}
\footnotetext{\textsuperscript{$\ddagger$}These authors jointly supervised this work.}

\let\thefootnote\oldthefootnote 
\makeatother
\endgroup

\begin{abstract}
Scientific discovery is driven by the iterative process of background research, hypothesis generation, experimentation, and data analysis. Despite recent advancements in applying artificial intelligence to scientific discovery, no system has yet automated all of these stages in a single workflow. Here, we introduce Robin, the first multi-agent system capable of fully automating the key intellectual steps of the scientific process. By integrating literature search agents with data analysis agents, Robin can generate hypotheses, propose experiments, interpret experimental results, and generate updated hypotheses, achieving a semi-autonomous approach to scientific discovery. By applying this system, we were able to identify a novel treatment for dry age-related macular degeneration (dAMD), the major cause of blindness in the developed world. Robin proposed enhancing retinal pigment epithelium phagocytosis as a therapeutic strategy, and identified and validated a promising therapeutic candidate, ripasudil. Ripasudil is a clinically-used rho kinase (ROCK) inhibitor that has never previously been proposed for treating dAMD. To elucidate the mechanism of ripasudil-induced upregulation of phagocytosis, Robin then proposed and analyzed a follow-up RNA-seq experiment, which revealed upregulation of \textit{ABCA1}, a critical lipid efflux pump and possible novel target. All hypotheses, experimental plans, data analyses, and data figures in the main text of this report were produced by Robin. As the first AI system to autonomously discover and validate a novel therapeutic candidate within an iterative lab-in-the-loop framework, Robin establishes a new paradigm for AI-driven scientific discovery.
\end{abstract}


\section{Introduction}
Advances in our ability to measure, perturb, and model biological systems have resulted in rapid growth of our collective scientific knowledge~\cite{stephens_big_2015}. Yet, complementary technologies to interpret, synthesize, and generate hypotheses from this knowledge have lagged behind~\cite{nurse_biology_2021}. Artificial intelligence (AI) systems based on large language models (LLMs) show promise for automating this knowledge synthesis process and accelerating scientific discovery. As a primary goal of biomedical research is the development of new treatments for disease, our ability to produce new therapeutics may be the ultimate beneficiary of these approaches. Drug development heavily relies on a confluence of biological, clinical, and pharmaceutical expertise, and is limited by the rate at which these experts can synthesize the scientific literature~\cite{cheng_importance_2020}.

The repurposing of existing drugs for new indications presents a promising application space for LLM systems. The history of drug repurposing often shows a pattern: while insights often existed in scientific literature, only after a significant lag did that knowledge crystallize into a new treatment. For example, dabrafenib, an inhibitor of BRAF kinase that is used in various cancers with mitogenic mutations in BRAF, is being repurposed to prevent hearing loss~\cite{ingersoll_braf_2020, ingersoll_dabrafenib_nodate}. While its molecular action was well characterized by 2010~\cite{noauthor_clinical_nodate, glaxosmithkline_phase_2017, ribas_braf_2011}, dabrafenib's otoprotective effects were only discovered 10 years later via unbiased high-throughput screening~\cite{noauthor_clinical_nodate}. This delayed discovery occurred despite dabrafenib’s otoprotective effects being a direct result of its known inhibition of BRAF~\cite{ingersoll_braf_2020, lahne_damage-induced_2008, ingersoll_dabrafenib_nodate}, suggesting that more repurposing opportunities could be identified through logical connection of existing biological insights in the literature. Further cases across medicine – from ketamine (22 year lag~\cite{berman_antidepressant_2000, noauthor_pcp_nodate}) to leucovorin (5 year lag~\cite{frye_cerebral_2013, ramaekers_folate_2008}) to KarXT (13 year lag~\cite{sauerberg_novel_1992, noauthor_krtx-10k_20191231htm_nodate}) – underscore how repurposing efforts are frequently realized years after core insights are documented.

Such delays in connecting existing insights to new therapeutic applications highlight the challenge of synthesizing disparate scientific knowledge. Trained on data across many fields, large language models (LLMs) are able to store and recall information on a wide variety of scientific topics and thus transcend the limitations of individual human knowledge. Previous work has shown that fine-tuned LLMs and specialized RAG systems can exceed human performance on retrieving and summarizing information from the scientific literature. These advances raise the possibility that LLM systems could be used for novel hypothesis generation~\cite{taylor_galactica_2022, giglou_llms4synthesis_2024, agarwal_litllm_2025, wang_using_2024}.

Several LLM systems have recently been developed to automate hypothesis generation~\cite{huang_automated_2025, boiko_autonomous_2023, lu_ai_2024, ifargan_autonomous_2024, wang_txgemma_2025, gottweis_towards_2025}. Specialized systems have also been developed to automate specific tasks in drug discovery, such as prediction of pharmacological properties and safety profiles~\cite{gottweis_towards_2025, lu_ai_2024}. These systems have demonstrated they can generate reasonable hypotheses by utilizing multi-agent architectures that decompose scientific reasoning into discrete manageable sub-tasks~\cite{gottweis_towards_2025}, domain-specific fine-tuning~\cite{wang_txgemma_2025, chaves_tx-llm_2024}, integration of external tools~\cite{wang_txgemma_2025, gottweis_towards_2025}, and incorporation of human feedback~\cite{gottweis_towards_2025}. However, none of these systems are currently capable of fully automating the key intellectual steps of the scientific process, including generating hypotheses and experimental strategies, analyzing results from the experiments, and refining hypotheses in light of new data.

Here, we introduce Robin, the first multi-agent system for scientific discovery that integrates novel hypothesis generation with experimental data analysis in one continuous workflow (Figure 1A). Robin utilizes specialized language agents for literature search (Crow and Falcon) and data analysis (Finch) to enable semi-autonomous scientific discovery~\cite{skarlinski_language_2024, Mitchener_Laurent_Tenmann_Narayanan_Wellawatte_White_Sani_Rodriques_2025}. Though this system could be applicable to scientific discovery across disciplines, in this report, we focus on its potential in therapeutics. Given a disease, Robin automatically identifies relevant \emph{in vitro} assays that model key disease mechanisms and proposes specific drug candidates to evaluate in these experimental models. Researchers next conduct the experiments and provide the resulting data to Robin for autonomous analysis. Robin then interprets the results of this analysis to generate a new round of therapeutic candidates. Through this process, Robin drives an iterative therapeutics development cycle where hypotheses are generated, tested, analyzed, and refined based on experimental results. The key intellectual steps of the scientific method are thus automated while coordinating with scientists throughout the experimental loop. By connecting literature-based hypothesis generation with experimental data analysis in a continuous feedback system, Robin represents the first complete implementation of AI-driven scientific discovery. 

To demonstrate Robin’s ability to generate and refine novel therapeutic hypotheses, we attempted to identify potential new treatments for dry age-related macular degeneration (dAMD). dAMD is the leading cause of irreversible sight loss in developed countries, yet limited treatment options are available. In the U.S. alone, 1.5 million people have vision-threatening dAMD, and 600,000 are legally blind due to AMD, a figure projected to almost triple by 2050 due to an aging population~\cite{Fleckenstein_Schmitz-Valckenberg_Chakravarthy_2024, rein_forecasting_2009}. By applying Robin to discover novel therapeutic candidates for dAMD, this work represents a first step towards AI-generated discovery in scientific research.

\begin{figure}[t]
\includegraphics[width=\columnwidth]{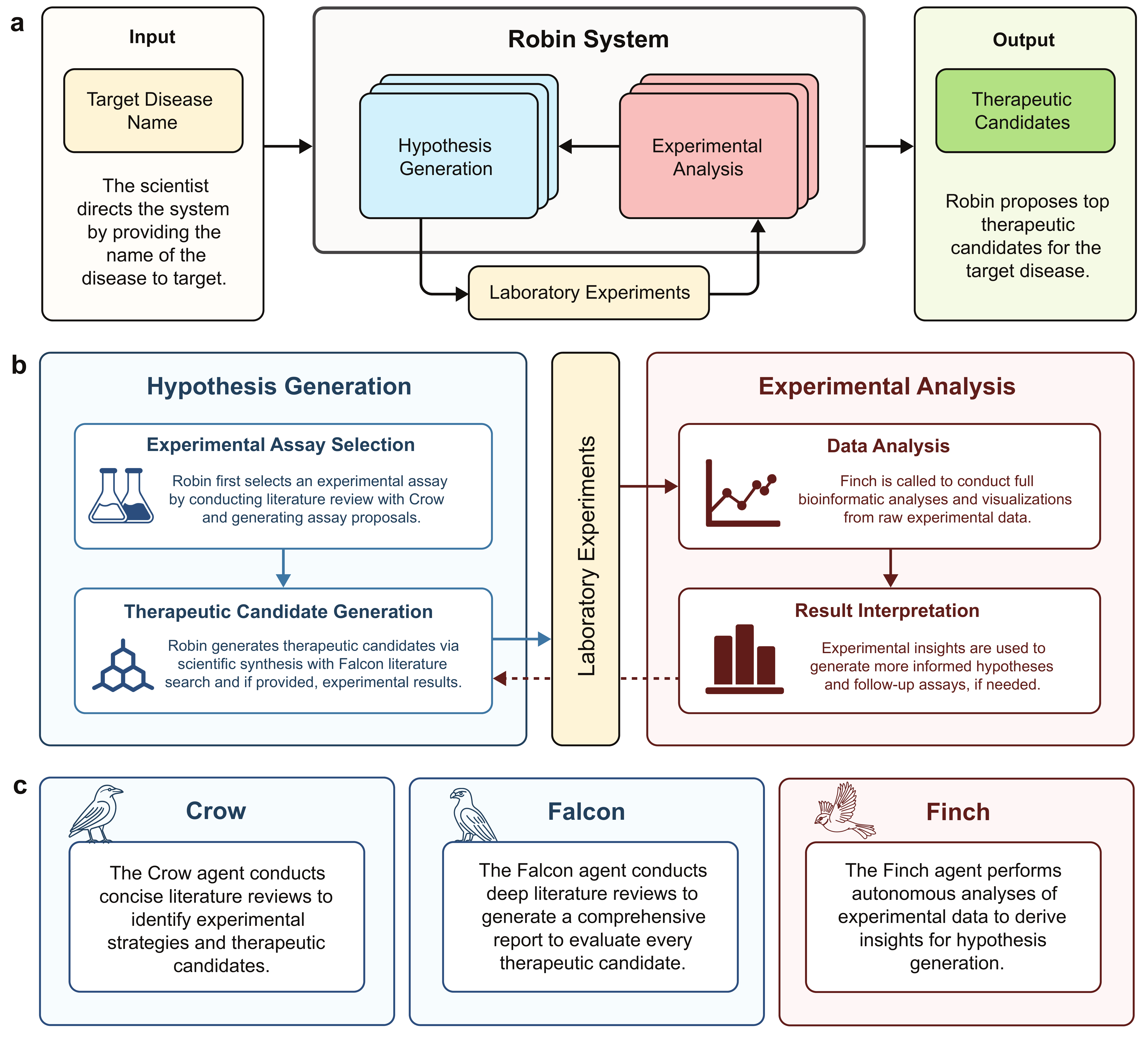}
\caption{\textbf{Architecture and workflow of the Robin system.} A) Given the name of a target disease, Robin generates hypotheses and selects top therapeutic candidates to test experimentally. Robin can autonomously analyze raw data from these experiments to synthesize scientific insights and generate updated therapeutic hypotheses. B) Robin interacts with language agents to generate hypotheses and analyze experimental data. C) Crow and Falcon are used to conduct concise and deep literature searches, respectively, to gather information to guide hypothesis generation. Finch performs analyses of experimental data, which Robin uses to derive insights to inform the next round of hypothesis generation.}
\end{figure}

\section{Results}
\subsection{Robin: A multi-agent system for scientific discovery}

Robin integrates multiple language agents in a structured workflow to generate therapeutic candidates for a given disease (Figure 1A,B). Crow and Falcon are literature search agents based on PaperQA2 that conduct concise and deep literature summaries, respectively~\cite{skarlinski_language_2024}. PaperQA2 achieves expert-level performance in information retrieval and summarization, with access to  scientific literature, clinical trial reports, and the Open Targets Platform~\cite{10.1093/nar/gkae1128}. Finch is a scientific data analysis agent that performs analyses of experimental data from assays, such as RNA-seq and flow cytometry (Figure 1C)~\cite{Mitchener_Laurent_Tenmann_Narayanan_Wellawatte_White_Sani_Rodriques_2025}. By coordinating these agents to identify novel therapeutics, Robin enables an experimentally-guided system that drives the process of scientific discovery. An example of the Robin hypothesis generation workflow is shown in Supplementary Figure \ref{fig:robinworkflow}.

\subsubsection{Therapeutic hypothesis generation}

When provided with a disease name, Robin formulates a series of general questions about the disease pathology and queries Crow to answer each question (Supplementary section \ref{sec:prompts_used}). Using the reports from Crow as context, Robin next identifies 10 potential causal disease mechanisms. For each mechanism, Robin again deploys Crow to prepare a detailed report describing an \emph{in vitro} model of the disease mechanism and corresponding assay that can be used to test drug efficacy. Robin uses an LLM judge to make pairwise comparisons between reports, which are used to calculate their relative rankings (see Methods). The top-ranked \emph{in vitro} model is used by Robin to define the experimental strategy for therapeutic candidate hypothesis generation.

Once an \emph{in vitro} model is selected, Robin conducts a similar sequence of general literature review and hypothesis generation to propose 30 therapeutic candidates for experimental testing. Robin then queries Falcon to generate a detailed report to evaluate each candidate. These reports contain both justification for why each drug is suitable for mitigating the disease mechanism represented in the \emph{in vitro} model and potential limitations the drug may pose. The drug candidates are ranked by an LLM-judged tournament according to the strength of the scientific rationale, pharmacological profile, and methodology of the supporting literature. This ranked list can then be reviewed by human scientists, and top drug candidates can be tested in the lab by executing an experimental protocol based on the assay suggested by Robin.

\begin{figure}[t]
\includegraphics[width=\columnwidth]{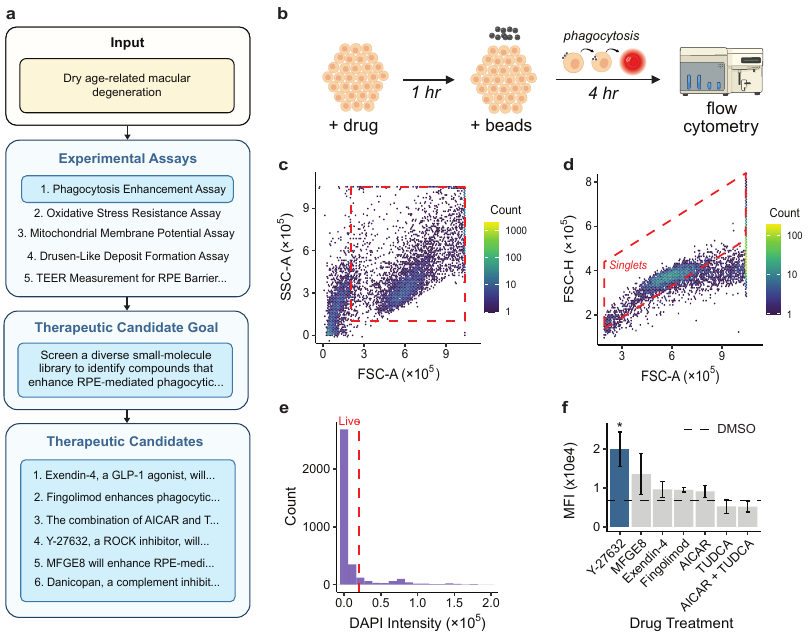}
\caption{\textbf{Robin generates therapeutic candidate hypotheses for dry AMD and analyzes experimental data from \textit{in vitro} tests} A) Robin proposes several experimental assays and ultimately decides to use an RPE phagocytosis enhancement assay. Robin synthesizes this strategy into an overall goal and then generates several novel therapeutic candidates to enhance RPE phagocytosis. B) Schematic representation of the phagocytosis assay. RPE cells are incubated with the drug for 1 hour before pHrodo beads are added. The cells are incubated with the beads for 3 hours and phagocytic activity is measured via flow cytometry. C-F) Example plots from a Finch flow cytometry analysis trajectory, formatted for readability in publication by a human. C) Finch performs gating to discard debris using a FSC-A vs SSC-A plot. D) Finch gates singlet cells from the FSC-H vs FSC-A plot. E) Finch identifies the DAPI signal and excludes dead cells. F) Finch performs statistical tests to compare candidate drugs to the DMSO control and plots the results.}
\end{figure}

\subsubsection{Experimental data analysis}

Once experiments are complete, the scientist uploads raw or semi-processed data and prompts Robin with a desired analysis approach, e.g., “RNA-seq differential expression analysis” or “flow cytometry". Robin then deploys Finch to carry out the desired analysis. This analysis step presents unique challenges due to the inherently ambiguous nature of biological data interpretation. For example, the gating choices in flow cytometry analysis or the filters used in RNA-seq analyses will vary between human scientists and may impact the final conclusions. Similarly, Finch's analysis results can vary between runs, even when given identical prompts and data, due to the stochasticity of the language agent. To leverage this diversity, Robin can launch 10 Finch analysis trajectories, each of which independently analyzes the experimental data. In each trajectory, Finch executes analysis code in a Jupyter notebook and provides an interpretable and reproducible summary of its findings. After all trajectories are complete, a meta-analysis can be conducted to synthesize all outputs into a consensus-driven conclusion. In this way, Finch both explores diverse analytical trajectories while delivering highly consistent end results based on consensus (Methods)~\cite{Narayanan_Braza_Griffiths_Ponnapati_Bou_Laurent_Kabeli_Wellawatte_Cox_Rodriques_et}.  

Finally, Robin distills actionable scientific insights from these processed experimental results. Robin also has the ability to propose targeted follow-up assays to explore or confirm significant or unexpected findings. These experimental insights are used to inform the next cycle of therapeutic hypothesis generation. The cycle continues until a human has a satisfactory novel drug candidate.

\subsection{Robin identifies novel therapeutic candidates for dry AMD}

We applied Robin to generate therapeutic candidates for dAMD as an initial proof-of-concept (Figure 2A). Robin began the therapeutic hypothesis generation workflow by identifying and reviewing 151 papers to propose ten biologically relevant dAMD mechanisms to assay. After ranking the disease mechanisms and corresponding experimental strategies, Robin proposed treating dAMD by increasing RPE cell phagocytosis, and suggested testing how well drugs increase the phagocytic capacity of either patient-derived RPE cells or ARPE-19 cells in a flow cytometry assay. 

Robin then deployed Crow to conduct a literature review of about 400 papers about RPE phagocytosis and the therapeutic landscape of dry AMD and synthesized the results to propose 30 existing drug candidates for experimental testing in the phagocytosis assay. Robin called Falcon to produce comprehensive evaluation reports on each of these molecules (Supplementary Figure \ref{fig:prompt_detailed_investigation_thera_cand}), which were ranked in an LLM-judged tournament. 

\begin{figure}[t!]
\includegraphics[width=\columnwidth]{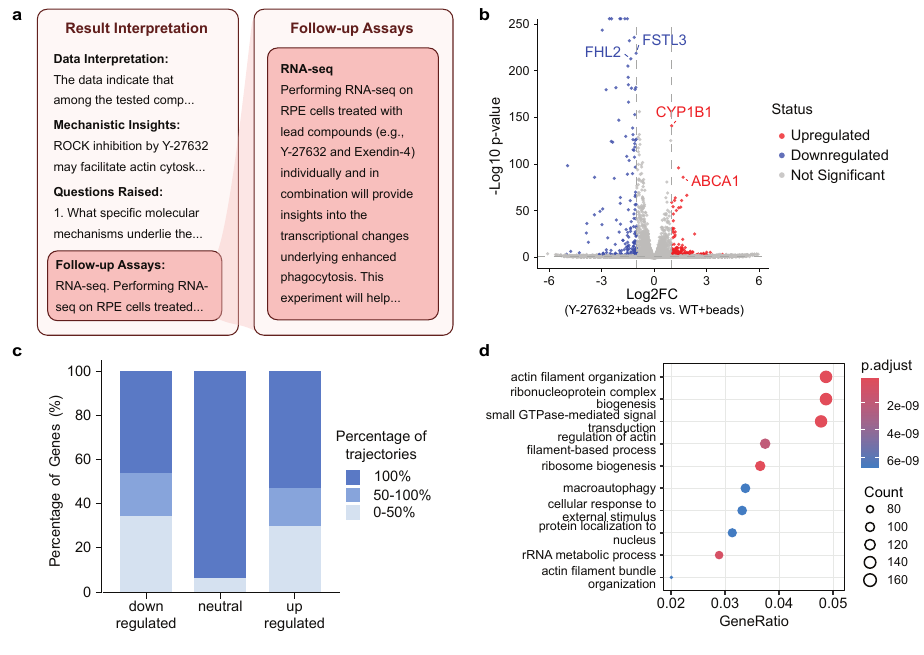}
\caption{\textbf{RNA-sequencing analysis of ARPE-19 cells treated with ROCK inhibitor Y-27632.} A) Robin interprets results from the first experiment and proposes follow-up assays. B-D) Example plots from a Finch RNA-seq analysis, formatted for readability in publication by a human. B) Finch-made volcano plot showing differentially expressed genes between Y27632-treated and wildtype cells after phagocytosis. C) Finch-made consensus findings from eight RNA-seq analysis trajectories, showing the percentage of analyses that identified the same genes as consistently up- or down-regulated. D) Finch-made GO-term enrichment of differentially expressed genes.}
\end{figure}

\subsubsection{ROCK inhibitors enhance RPE phagocytosis}

We then selected the top five candidates from this ranking for experimental testing: Exendin-4, Fingolimod, MFGE8, Y-27632, and the combination of AICAR and TUDCA. Although Robin suggested using fluorescently labeled photoreceptor outer segments as a substrate for RPE phagocytosis, we instead decided to use pHrodo beads due to availability. pHrodo beads are fluorescently activated in the low pH environment of the lysosome, allowing detection of phagocytosis in single cells using flow cytometry (Figure 2B). After testing these therapeutic candidates in the RPE phagocytosis assay, the raw flow cytometry data, associated metadata, and an analysis prompt were uploaded to Robin. Robin called the data analysis agent Finch, which developed a Jupyter notebook to quantify the effect of each compound on RPE phagocytosis by gating the flow cytometry data and performing statistical tests. Plots from a representative individual Finch trajectory are shown in Figure 2C-F. These results were confirmed by a human analysis of the same data (Supplementary Figure \ref{fig:sup_flow_cytometry_round1}).  Preclinical models have demonstrated that Y-27632 can restore phagocytic efficiency in RPE cells~\cite{Mao_Finnemann_2012}, confirming Robin's literature-based rationale for suggesting this candidate.

\subsubsection{Automated RNA-seq analysis characterizes transcriptional changes induced by ROCKi-enhanced phagocytosis}

After analyzing the initial flow cytometry results, Robin recommended RNA sequencing of Y-27632-treated RPE cells to investigate the transcriptional effects of ROCK inhibition (Figure 3A). We conducted a second RPE phagocytosis experiment with Y-27632 and profiled the samples using bulk RNA sequencing. Finch conducted differential gene expression (DGE) analysis and summarized results in a volcano plot (Figure 3B). While previous studies have shown that Y-27632 enhances phagocytic cup formation through post-translational regulation of F-actin dynamics~\cite{Mao_Finnemann_2021, Halasz_Townes-Anderson_2016, Hollanders_VanBergen_Kindt_Castermans_Leysen_Vandewalle_Moons_Stalmans_2015, Kamao_Miki_Kiryu_2019, Maekawa_Ishizaki_Boku_Watanabe_Fujita_Iwamatsu_Obinata_Ohashi_Mizuno_Narumiya_1999}, Finch's DGE analysis revealed that Y-27632 treatment also induced rapid transcriptional changes in RPE cells during phagocytosis. Finch's consensus analysis demonstrated that Finch identified the same genes as significantly differentially expressed in over 50\% of trajectories (Figure 3C). Finch next performed GO enrichment analysis and found that Y-27632 significantly altered the expression of genes involved in actin filament organization, small GTPase-mediated signal transduction, and autophagy pathways upon phagocytosis (Figure 3D). These results suggest that Y-27632 enhances the initial uptake phase of phagocytosis through cytoskeletal rearrangements and promotes clearance of internalized material through transcriptional regulation of autophagy. Further work is necessary to confirm whether these changes are specific to Y-27632 or are generalizable to any intervention that significantly increases phagocytic capacity. The human version of this analysis is similar to that of Finch and is shown in (Supplementary Figure \ref{fig:human_RNAseq})

The DGE analysis identified a 3-fold upregulation (adjusted p=2.13x10$^{-83}$) of \emph{ABCA1}, a critical lipid efflux pump, in Y-27632-treated cells. Differential expression of \emph{ABCA1} upon ROCKi-induced phagocytosis has significant implications for dAMD, as \emph{ABCA1} is essential for healthy RPE function. \emph{ABCA1} facilitates the active transport of cholesterol and phospholipids from the plasma membrane to acceptor proteins prior to their ejection from the cell. Notably, \emph{ABCA1} belongs to the same transporter family as \textit{ABCA4}, which has been previously identified as a therapeutic target in macular degeneration pathogenesis \cite{Al-Khuzaei_Broadgate_Foster_Shah_Yu_Downes_Halford_2021}. Further connecting these findings to dAMD pathology, Apo-E, a lipid acceptor for \emph{ABCA1}, has also been identified as a potential therapeutic target for dAMD \cite{Malek_Johnson_Mace_Saloupis_Schmechel_Rickman_Toth_Sullivan_Rickman_2005}. These mechanistic insights, derived from experiments proposed by Robin and analyzed by Finch, demonstrate how AI-driven scientific discovery can not only identify effective therapeutic compounds but also reveal novel molecular targets within disease pathways that might have otherwise remained unexplored. 

\begin{figure}[ht]
\includegraphics[width=\columnwidth]{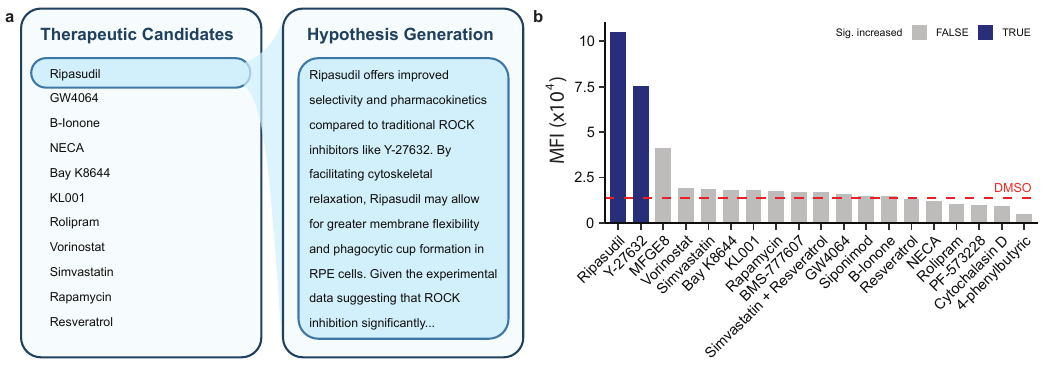}
\caption{\textbf{Ripasudil significantly enhances RPE phagocytosis.} A) Excerpt of Robin proposal for ripasudil. Drawing from the insights from the first round of experimental analysis, Robin proposes ripasudil as a therapeutic candidate for treating dry AMD. B) Analyzed flow cytometry data from the second round of experiments shows that ripasudil significantly enhances phagocytosis in RPE cells, inducing an even greater effect than Y-27632.}
\end{figure}

\subsubsection{Ripasudil as a drug repurposing candidate for dry AMD}

In addition to suggesting RNA-seq analysis on Y-27632, Robin also conducted a subsequent iteration of candidate drug hypotheses. We tested 10 of these drugs experimentally and provided the data to Finch for analysis. Finch’s analysis showed that ripasudil, a ROCK inhibitor approved for treatment of glaucoma in Japan, outperformed Y-27632 and increased RPE cell phagocytosis 7.5-fold compared to DMSO controls (Figure 4A,B; human analysis showed 1.75-fold increase in Supplementary Figure \ref{fig:sup_flow_cytometry_round2}). Although further testing at different doses and longer incubation times would be necessary for definitive comparison, the initial superior performance of ripasudil relative to Y-27632 demonstrates Robin's ability to progressively refine therapeutic hypotheses through iterative experimentation and feedback.

\section{Discussion}

In this report, we present Robin, a multi-agent system integrating automated hypothesis generation and experimental data analysis for scientific discovery. When tasked with identifying novel therapeutics for dAMD, Robin proposed enhancing RPE cell phagocytosis using ROCK inhibitors, and discovered ripasudil as the most potent enhancer of RPE phagocytosis among tested compounds through an iterative lab-in-the-loop discovery cycle. Ripasudil's established safety profile and clinical approval for ocular use present a promising drug repurposing opportunity that could significantly accelerate the development pathway for dAMD treatment.

Notably, while ROCK inhibitors have been previously suggested for treatment of wet AMD and other retinal diseases of neovascularization, Robin is the first to propose their application in dry AMD for their effect on phagocytosis~\cite{Halasz_Townes-Anderson_2016}. This approach is supported by several lines of evidence, as RPE phagocytic dysfunction is observed not only with normal aging but is pronounced in AMD patients~\cite{inana_rpe_2018, si_role_2023, kaarniranta_autophagy_2013}. In investigating the basis for Robin’s initial identification of Y-27632, we found that the literature search included a single paper demonstrating Y-27632’s ability to enhance phagocytic efficiency in “low-phagocytic” RPE cells from human donors by promoting actin polymerization~\cite{si_role_2023}. Taken together, these results demonstrate Robin's ability to effectively generate novel hypotheses by synthesizing insights already present in the scientific literature.

Beyond the specific application of dAMD, Robin addresses a broader challenge in therapeutic development. With FDA approvals stagnating at approximately 50 novel drugs annually over the past decade~\cite{baedeker_2024_2025}, new approaches to scale therapeutic discovery are urgently needed. As the first system to automate both literature-grounded hypothesis generation and experimental data analysis, Robin is poised to accelerate the pace of drug discovery compared to traditional approaches. We demonstrate Robin's broader applicability by generating hypotheses for 10 additional diseases with pressing therapeutic needs (Supplementary Figures \ref{fig:pcos}-\ref{fig:ckd}). 

In its current implementation, Robin has several opportunities for continued development. For example, while Robin generates experimental outlines, it does not yet produce precise, executable protocols—future iterations aim to provide detailed methodologies that require minimal human translation for laboratory execution. The Finch data analysis agent is also heavily reliant on prompt engineering by domain experts to produce reliable analytical results. Adapting Finch to independently generate or adapt prompts to specific data modalities would enable a more autonomous discovery pipeline. Finally, while Robin uses an LLM-judged tournament to nominate therapeutic hypotheses, future work on better aligning hypothesis generation and evaluation with human scientific judgment may be helpful in more reliably producing high-quality hypotheses. 

By automating hypothesis generation, experimental planning, and data analysis in an integrated system, Robin represents a powerful new paradigm for AI-driven scientific discovery. This approach can be used to not only reshape therapeutic development, but fundamentally accelerate the scientific process to drive a greater understanding of the natural world.

\section{Methods}

\subsection{Robin Implementation}
Robin is implemented as a Jupyter notebook using the Aviary framework to instantiate and call agents~\cite{Narayanan_Braza_Griffiths_Ponnapati_Bou_Laurent_Kabeli_Wellawatte_Cox_Rodriques_et}. Robin utilizes the OpenAI o4-mini model to synthesize literature and generate hypotheses, and the Anthropic Claude 3.7 Sonnet model as the judge for pairwise comparisons to rank hypotheses.

Notably, while the experiments in this paper were conducted using an agentic implementation of Robin, we observed that Robin almost always called tools in the same order, leading to a deterministic workflow. Therefore, we translated Robin into a streamlined Jupyter notebook to improve stability and ease of use.

The ranking of hypotheses was calculated via a series of pairwise comparisons adjudicated by the LLM judge. When ranking 25 or fewer hypotheses, all possible pairs were compared in a round robin tournament. When considering more than 25 hypotheses, 300 pairwise comparisons were randomly selected to achieve a comprehensive assessment within reasonable computational and time constraints. The outcomes from these comparisons were used to estimate strength parameters via the Bradley-Terry-Luce (BTL) model, which subsequently informed the relative ranking of each hypothesis~\cite{bradley_rank_1952}.

To generate the prompt for this LLM judge, domain experts were asked to conduct pairwise comparisons of Robin’s therapeutic candidate hypotheses. The results of these expert evaluations were given to Google's Gemini 2.5 Pro Preview model to generate a prompt for Robin’s LLM judge. This approach was designed to elicit decision-making from the LLM judge consistent with expert preferences and criteria.

When comparing the LLM judge’s preferences with experts’, the LLM judge demonstrated high concordance with expert preferences, with an average of 7.25 of its top 10 hypotheses matching those in the experts' top 10 (Supplementary Figure \ref{fig:humanevals}A). This concordance is more than double that expected from random selection. Furthermore, the LLM judge exhibited higher intra-rater consistency than human experts (Supplementary Figure \ref{fig:humanevals}B). When presented with identical pairwise comparisons, the LLM judge selected the same hypothesis in 88\% of comparisons, as compared with human experts who selected the same hypothesis 61\% of the time.

\subsection{Finch Implementation}

Finch, initially introduced in BixBench, is an autonomous, Jupyter-native data analysis agent designed using the Aviary framework \cite{Narayanan_Braza_Griffiths_Ponnapati_Bou_Laurent_Kabeli_Wellawatte_Cox_Rodriques_et}. Finch systematically processes bioinformatics workflows, such as RNA sequencing differential expression analysis or flow cytometry, based on a provided dataset and research question.

The agent operates within a structured execution environment provided by the Aviary framework, an extensible gymnasium tailored for language agent evaluation and iterative problem-solving. Aviary was selected specifically for its controlled environment, supporting reproducible evaluations through standardized software environments, consistent tool access, and structured multi-step reasoning.

Finch leverages an agentic prompting strategy based on the ReAct approach, balancing logical reasoning and practical execution effectively \cite{Yao_Zhao_Yu_Du_Shafran_Narasimhan_Cao_2023}. Each Finch trajectory unfolds within a pre-built Docker container (BixBench-env:v1.0), containing extensive bioinformatics-oriented Python, R, and Bash libraries. This standardized container ensures reproducibility and isolates evaluation strictly to Finch’s analytical capability rather than software installation or dependency management. That being said, Finch is capable of installing new dependencies when needed and does so competently.

\subsubsection{Tools and Environment}

Finch interacts exclusively via two tools enabled by the Aviary framework:
\begin{itemize}
\item \texttt{edit\_cell}: The agent can select, modify, and execute cells within a Jupyter notebook.
\item \texttt{submit\_answer}: Finalizes and submits the agent's analytical conclusion.
\end{itemize}

Prompt engineering was thoroughly explored to optimize Finch’s initial instructions, with an example provided in Supplementary Figure \ref{fig:finch_prompt}. All Finch trajectories are available through the FutureHouse platform.

\subsection{Cell Culture}

ARPE-19 cells (ATCC; Cat. No. CRL-2302) are maintained in DMEM/F12 (Corning; Cat. No. 15-090-CV), with 10\% fetal bovine serum (Gibco; Cat. No. 10082-147), 2mM L-glutamine (Gibco; Cat. No. 25030-081), 1mM sodium pyruvate (Gibco; Cat. No. 11360-070) and 1\% penicillin-streptomycin (Gibco; Cat. No. 15140-122) at 37°C, 5\% CO$_{2}$. For phagocytosis assays, cells are seeded in 96-well tissue culture plates at a density of 1x$10^{4}$ cells per well with 100 $\mu$L of complete cell media. They are grown to confluence and then incubated for a further 7 days, during which the media is not changed. At this point they have formed a monolayer and show signs of differentiation (melanin granule expression, cobblestone morphology)~\cite{DUNN_AOTAKI-KEEN_PUTKEY_HJELMELAND_1996}. 

\subsection{Phagocytosis Assay}

\subsubsection{Drug Library Preparation}

Optimal cell culture concentrations for the drugs proposed by Robin were found in the literature and if multiple concentrations were proposed, the highest was selected for use in the phagocytosis assay. On the day of the phagocytosis experiment, drug stocks were used to prepare working stocks in complete media at 2X working concentration. AICAR and TUDCA were used in combination at the concentrations below.

\begin{table}[htbp]
\centering
\label{tab:drugs_concentrations}

\renewcommand{\arraystretch}{1.2} 

\begin{tabularx}{\columnwidth}{| >{\raggedright\arraybackslash}X | l |}
\hline
\textbf{Drug} & \textbf{Working concentration} \\
\hline

Y-27632 (Cayman Chemical Company; Cat. No. 10005583) & 20$\mu$M~\cite{sun_y-27632_2015} \\
\hline
AICAR (Sigma-Aldrich; Cat. No. A9978) & 1mM~\cite{chung_aicar_2017} \\
\hline
Tauroursodeoxycholic acid (TUDCA) (TargetMol; Cat. No. T2532) & 100$\mu$M~\cite{li_tudca_2019} \\
\hline
Exendin-4 (TargetMol; Cat. No. T3967) & 1$\mu$M~\cite{cui_exendin-4_2019} \\
\hline
MLN120$\beta$ (TargetMol; Cat. No. TQ0306) & 20$\mu$M~\cite{Hideshima_Neri_Tassone_Yasui_Ishitsuka_Raje_Chauhan_Podar_Mitsiades_Dang_et} \\
\hline
Fingolimod (PeproTech; Cat. No. 1625605) & 1$\mu$M~\cite{zhou_immunomodulatory_2009} \\
\hline
MFGE-8 (Sino Biological; Cat. No. 10853-H27B-B-20) & 2.5$\mu$g/mL~\cite{Chiang_Chen_Lin_Tsai_2008} \\
\hline
Ripasudil (Cayman Chemical Company; Cat. No. 19920) & 100$\mu$M~\cite{yang_ripasudil_2024} \\
\hline
Vorinostat (Active Motif; Cat No. 14027) & 5$\mu$M~\cite{maksimova_histone_2023} \\
\hline
Simvastatin (TargetMol; Cat No. T0687) & 1$\mu$M \\
\hline
Resveratrol (PeproTech; Cat No. 5013606) & 16mM~\cite{wagner_ascorbic_2011} \\
\hline
Bay K8644 (Alomone Labs; Cat No. B-350) & 3$\mu$M~\cite{lei_role_2008} \\
\hline
KL001 (Cayman Chemical Company; Cat No. 13878) & 50$\mu$M~\cite{hirota_identification_2012} \\
\hline
Rapamycin (Gift from RetroBio) & 100nM~\cite{semlali_rapamycin_2022} \\
\hline
BMS-777607 (ApexBio; Cat No. A5703) & 10$\mu$M~\cite{torka_activation_2014} \\
\hline
NECA (5’-N-ethylcarboxamidoadenosine; Cayman Chemical Company; Cat No. 21420) & 10$\mu$M~\cite{ohta_vitro_2009} \\
\hline
Rolipram (PeproTech; Cat No. 6145453) & 10$\mu$M \\
\hline
PF-573228 (ApexBio; Cat No. B1523) & 5$\mu$M~\cite{aulakh_inhibiting_2018} \\
\hline
4-phenylbutyric acid (ApexBio; Cat No. C6831) & 10mM~\cite{jiang_4-phenylbutyric_2021} \\
\hline
B-Ionone (AA Blocks; Cat No. AA003JVY) & 200$\mu$M~\cite{kang_-ionone_2013} \\
\hline
Siponimod (TargetMol; Cat No. T6403) & 10mM~\cite{sartawi_bone_2020} \\
\hline
\end{tabularx}

\caption{List of Drugs and Working Concentrations}
\end{table}

\subsubsection{pHrodo Bead Preparation}

To specifically detect phagocytosed particles, we utilized pHrodo beads (ThermoFisher; Cat. No. A10010) which are fluorescent only in the low pH of the lysosome. Deep red \emph{E. coli} pHrodo beads were thawed and a single vial was resuspended in 2 ml PBS as per the manufacturer's instructions. The resuspended beads (protected from light) were then sonicated for 10 minutes in a water bath at RT before addition to the cell assay plate. 

\subsubsection{Phagocytosis Assay}

For the phagocytosis assays, cells were treated with test compounds or vehicle control (0.5\% DMSO) for 60 minutes at 37\textdegree C, 5\% CO$_{2}$, prior to bead addition by adding 100 $\mu$L of 2X drug stock in complete media to the cells. At 60 minutes, 10 $\mu$L of pHrodo beads (approximately 10$\mu$g of beads) were added to each well. After bead addition, plates were incubated at 37\textdegree C for 3 hours while phagocytosis of the pHrodo beads proceeded. Cells were treated with TrypLE Express (Gibco; Cat. No. 12604-013) for 10 minutes and washed with FACS buffer (1\% BSA and 500 ng/ml DAPI in calcium and magnesium-free PBS) to achieve a cell suspension suitable for flow cytometry.

\subsubsection{Flow Cytometry}

ARPE-19 cells suspended in FACS buffer in 96-well cell culture plates were immediately transferred to an Attune NxT flow cytometer (ThermoFisher; Cat. No. A24858) equipped with an Autosampler. Fluorescence from the pHrodo beads and DAPI was stimulated using the 637 and 405 nm lasers, respectively, and detected using the 670/14 and 450/40 filters, respectively. All events were analyzed until the full volume of the well was aspirated to capture the entire cell population. A minimum of 4,000 events per well were recorded to ensure adequate statistical power (excluding the bead-only (no cell) control).

\subsubsection{RNA Sequencing}

ARPE-19 cells were seeded in 24-well tissue culture plates and cells were lysed and RNA extracted using the Maxwell RSC 48 (Promega; Cat. No. AS8500) and Maxwell RSC SimplyTissue kit (Promega; Cat. No. AS1340) as per the manufacturer’s guidelines. Total RNA was quantified using a Qubit instrument (ThermoFisher; Cat. No. Q33327) and Qubit high-sensitivity RNA kit (ThermoFisher; Cat. No. Q10210). Using 1 $\mu$g of total RNA as input, poly-A tailed mRNA was isolated using NEBNext® Poly(A) mRNA Magnetic Isolation Module (New England Biolabs; Cat. No. E7490) and then prepared for sequencing using the NEBNext Ultra II Directional RNA Library Prep Kit for Illumina (New England Biolabs; Cat. No. E7760). Quality control was performed using an Agilent Bioanalyzer. Libraries (with 1\% PhiX spiked in) were sequenced on an Illumina NextSeq 2000 with 75 base-pair paired-end reads using a P3 flow cell. 

\subsection{Data Analysis}

\subsubsection{Flow Cytometry}

Finch, the data analysis agent, performed end-to-end analysis of our flow cytometry data. Debris and cell aggregates were excluded from the analysis based on forward and side scatter parameters. In general, Finch clusters the forward scatter versus side scatter plot by using the flowMeans package for the R programming language to apply a k-means clustering before choosing the cluster most likely to contain the main cell population. For subsequent gates, Finch generated scatter plots and then used the plots to define gates. Single cells were gated based on pulse width and area parameters. The mean fluorescence intensity (MFI) of the pHrodo signal was quantified for each well to assess the extent of phagocytosis. 

Flow cytometry data was also analyzed by a human scientist. Those strategies are shown in Supplementary Figures 2 and 3. The human flow cytometry analysis was similar to Finch’s except that the human analysis had the additional step of using the no-bead control to remove background signal in the Alexa 647 channel, which proved minimal. The human analysis started by identifying the main cell population in the forward versus side scatter plot. Next, singlets were identified. Dead cells were then removed using the DAPI channel before finally removing Alexa 647 background signal using the no bead control. 

\subsubsection{RNA Sequencing}

Read demultiplexing and alignment was performed by a human and subsequent differential gene expression analysis was performed by Finch. 

\subsubsection{Read Alignment and Processing}

Raw paired-end RNA-Seq reads (2 × 150 bp) for twelve samples (including Y27632-treated, wild-type, and bead-treated conditions) were processed using a custom Bash pipeline. Sequencing files (*\_R1\_001.fastq.gz and *\_R2\_001.fastq.gz) were retrieved from the FASTQ directory and aligned individually to the GRCh38 human reference genome (HISAT2 index built from GENCODE v44) using HISAT2 with the --dta flag to retain splice junction information for downstream transcriptome assembly. Alignments were performed with eight CPU threads (-p 8) and output in SAM format. Each SAM file was converted to BAM with samtools view -bS, then sorted (samtools sort) and indexed (samtools index) to produce compressed, coordinate-sorted BAMs. Intermediate SAM and unsorted BAM files were removed to conserve disk space. This procedure ensured that all alignments were uniformly processed and ready for quantitative analysis.

\subsubsection{Gene-Level Quantification}

Gene counts were obtained with the featureCounts tool from the Subread package, using the GENCODE v44 GTF annotation file. Paired-end mode (-p) was specified to correctly handle fragment counting, and eight threads (-T 8) were used to accelerate processing. All of the indexed, sorted BAM files were supplied simultaneously, and read assignment to genes was performed according to the provided exon features. The resulting raw gene-level count matrix was written to gene\_counts.txt, serving as the input for subsequent normalization and differential expression analyses. This workflow provides a reproducible framework for high-throughput RNA-Seq data processing from raw reads through gene-level quantification.

\subsubsection{Differential Gene Expression Analysis}

Raw gene counts were imported into R (v4.2.0) and filtered to remove non-count columns, yielding a matrix of 6 samples across two conditions (Y-27632, untreated). The count matrix and metadata were encapsulated in a DESeqDataSet object (DESeq2 v1.36.0)~\cite{love_moderated_2014}. Normalization, dispersion estimation, and Wald tests were performed to detect differential expression. Transcript identifiers were mapped to gene symbols using biomaRt, and one pairwise contrasts were defined (Y-27632 versus untreated)~\cite{durinck_mapping_2009}. Results were ordered by adjusted p-value, exported to CSV, and volcano plots were generated with EnhancedVolcano (v1.14.0) using thresholds of |log$_{2}$FC| > 1 and adj. p < 0.05~\cite{noauthor_enhancedvolcano_nodate}.

\section*{Acknowledgements}

We would like to acknowledge Michael Skarlinski, Mayk Caldas, Tyler Nadolski, James Braza, and Siddharth Narayanan for their help in supporting this research and reviewing this manuscript. We are also grateful to all our colleagues at FutureHouse for their support. 

\section*{Data and Code Availability}

Sample trajectories for Robin and Finch, as well as the code for Robin will be available at \url{github.com/Future-House/robin}

\clearpage
\bibliographystyle{unsrt}  

\bibliography{references}

\newpage
\section{Supplementary Material}

\subsection{Supplementary Figures}

\begin{suppfigureenv}
\centering
\includegraphics[width=\columnwidth]{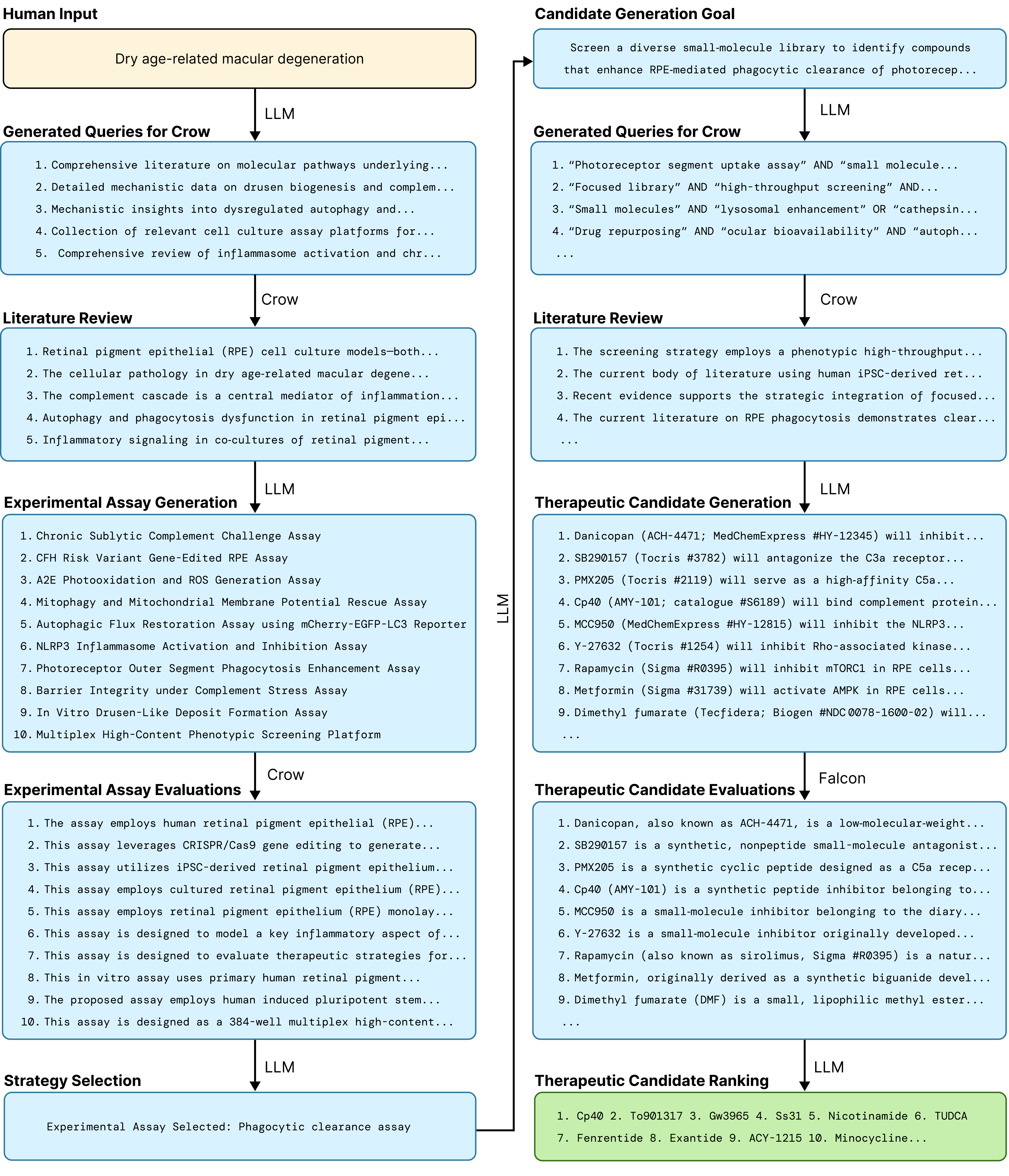}

\textbf{Supplementary Figure \thesuppfigure}{ Detailed example workflow of Robin experimental assay and therapeutic candidate hypothesis generation.}
\label{fig:robinworkflow}
\end{suppfigureenv}

\newpage

\subsection{Prompts used in this report}
\label{sec:prompts_used}

\subsubsection{Experimental Assay Generation}

\begin{suppfigureenv}
\centering
\begin{tcolorbox}[
    enhanced, 
    colback=black!5!white,   
    colframe=black!75!black, 
    arc=3mm,                
    boxrule=1pt,
]

\textbf{Model:} o4-mini

\tcbline 

\textbf{System prompt:}
\par 
You are a highly experienced biomedical research strategist. Your task is to propose **mechanistic hypotheses** tested via **cell-culture assays** or **therapeutic strategies**. Generate exactly \{num\_assays\} distinct ideas.

\tcbline 

\textbf{User prompt:}
\par 
Return a list of \{num\_queries\} queries (separated by <>) that would be useful in doing research to develop detailed, mechanistic cell culture assays that would be used to evaluate drugs to treat \{disease\_name\}.
These queries will be given to a 20-person scientific team to research in depth, so they should be able to capture broadly relevant information (30+ words) and search relevant literature across
biomedical, clinical, and biochemical literature about the disease or therapeutic landscape. Don't look up specific drugs, but any relevant scientific information that may help with assay development.
You have \{num\_queries\} queries, so spread your queries out to cover as much ground as possible. Create queries both about the general biochemistry and mechanistic underpinnings of \{disease\_name\} as well as about the assays.
In formatting, don't number the queries, just output a string with \{num\_queries\} queries separated by <>.

\end{tcolorbox}
\textbf{Supplementary Figure \thesuppfigure}{ Prompt for generating literature search queries.}
\label{fig:prompt_lit_search_queries}
\end{suppfigureenv}

\begin{suppfigureenv}
\centering
\begin{tcolorbox}[
    enhanced, 
    colback=black!5!white,   
    colframe=black!75!black, 
    arc=3mm,                
    boxrule=1pt,
]

\textbf{Model:} o4-mini

\tcbline 

\textbf{System prompt:}
\par 

You are a professional biomedical researcher, having experience in early-stage drug discovery and validation in vitro. Your task is to propose **mechanistic hypotheses** tested via **cell-culture assays** or **therapeutic strategies**. Generate exactly \{num\_assays\} distinct assay ideas. You are meticulous, creative, and scientifically rigorous. Focus on strategies that prioritize simplicity, speed of readout, biological relevance, and direct measurement of functional endpoints. Strong preference for biologically relevant strategies.

\quad **Output Format Specification (Strict Adherence Required):**

Your entire output MUST be a single, valid JSON object. This JSON object will be an **array** at its root, containing exactly \{num\_assays\} individual JSON objects. Each of these inner objects represents one distinct proposal and MUST conform to the following structure and content guidelines:

\begin{Verbatim}[breaklines]
[
{{
    "strategy_name": "string", # Name of the strategy. Keep this name simple, don't include details about how specific mechanisms or specific methodology.
    "reasoning": "string" # Scientific reasoning justifying the chosen strategy or the feasibility/relevance of the assay design, citing relevant literature.
}}
// ... more objects here, up to {num_assays} total
]
\end{Verbatim}

\tcbline 
\textbf{User prompt:}
\par 
Generate exactly **\{num\_assays\}** distinct and scientifically rigorous proposals for cell culture assays that can evaluate drugs to treat \{disease\_name\}. Here is some relevant background information that can guide your proposals: \{assay\_lit\_review\_output\}

\end{tcolorbox}
\textbf{Supplementary Figure \thesuppfigure}{ Prompt for generating hypotheses for experimental assays.}
\label{fig:prompt_exp_assay_hypotheses}
\end{suppfigureenv}

\clearpage

\begin{suppfigureenv}
\centering
\begin{tcolorbox}[
    enhanced, 
    colback=black!5!white,   
    colframe=black!75!black, 
    arc=3mm,                
    boxrule=1pt,
]
\textbf{Model:} Crow

\tcbline

You are a professional biomedical researcher, having experience in early-stage drug discovery and validation in vitro. Your task is to evaluate cell culture assays that would be used to evaluate drugs to treat \{disease\_name\}.
Given the following assay, do a comprehensive literature review to evaluate if this assay would be useful for testing therapeutics for \{disease\_name\}. Search relevant literature across biomedical, clinical, and biochemical literature about the disease or therapeutic landscape. Don't look up specific drugs, but any relevant scientific information that may inform assay development. \{Strategy\}

Provide your response in the following format, like an evaluation for a scientific proposal:

\quad \quad Assay Overview: Explain the assay idea, including the following key points: which aspect of the disease pathogenesis does the assay model, what measurements will be taken from the assay and how they will be taken, which cells or other biological material are used in the assay.

\quad \quad Biomedical Evidence: Make a compelling argument for how the aspect of the disease represented in the assay is central to the pathogenesis of the disease. Make sure to consider both the biomedical and clinical literature.

\quad \quad Previous Use: Explain how this assay has previously been used for drug discovery (if this has been done). Explain any key scientific discoveries which have been made using this assay.

\quad \quad Overall Evaluation: Strengths and weaknesses of this assay for testing therapeutics for \{disease\_name\}.

\end{tcolorbox}
\textbf{Supplementary Figure \thesuppfigure}{ Prompt for generating detailed evaluation for each experimental assay}
\label{fig:prompt_exp_assay_eval}
\end{suppfigureenv}

\begin{suppfigureenv}
\centering
\begin{tcolorbox}[
    enhanced, 
    colback=black!5!white,   
    colframe=black!75!black, 
    arc=3mm,                
    boxrule=1pt,
]

\textbf{Model:} Claude Sonnet 3.7

\tcbline

\textbf{System prompt:}
\par 
You are an experienced drug development committee member, with broad scientific expertise across the biology, chemistry, clinical, medical, and pharmaceutical sciences.

\quad Your objective is to perform a rigorous scientific comparison of two proposals for experimental assays that will be used to test therapeutics for \{disease\_name\}.

\quad Your evaluation must be based strictly on the presented scientific evidence, scientific novelty, methodological rigor, and logical interpretation, not on the persuasive quality or wording of the proposal documents.

\quad Critically assess the scientific soundness and biological rationale for the experimental assay, analyzing the supporting literature and historical usage.

\quad Preference for in vitro strategies that prioritize simplicity, speed of readout, biological relevance, and direct measurement of functional endpoints. Strong preference for biologically relevant strategies.

\quad The goal of this task is to choose the best in vitro experimental assay that would be scientifically relevant and insightful for testing therapeutics for \{disease\_name\}. Prefer assays that are biologically functionally relevant and simple to perform in standard lab setting.

\tcbline 

\textbf{User prompt:}
\par 
Evaluate the experimental assays using the structure below. This evaluation informs a critical decision.

\quad **Respond ONLY in the specified JSON format, do not include any text outside the JSON object itself.**

\begin{Verbatim}[breaklines]
{{
"Analysis": {"type": "string", "description": "[Provide a detailed analysis of the two experimental assays, based on the evaluation criteria and the evidence provided.]"},
"Reasoning": {"type": "string", "description": "[Choose which experimental assay is better. Provide a detailed explanation for why you think the winner is better than the loser, based on the evaluation criteria and the evidence provided.]"},
"Winner": {"type": "string", "description": "[Return the name and ID number of the candidate that you think is better between the two candidates, as a tuple. It should be formatted as (winner_name, winner_id)]"},
"Loser": {"type": "string", "description": "[Return the name and ID number of the candidate that you think is worse between the two candidates, as a tuple. It should be formatted as (loser_name, loser_id)]"},
}}
\end{Verbatim}

\end{tcolorbox}
\textbf{Supplementary Figure \thesuppfigure}{ Prompt for LLM judge to select the top experimental assay}
\label{fig:prompt_llm_judge_assay_selection}
\end{suppfigureenv}

\begin{suppfigureenv}
\centering
\begin{tcolorbox}[
    enhanced, 
    colback=black!5!white,   
    colframe=black!75!black, 
    arc=3mm,                
    boxrule=1pt,
]

\textbf{Model:} o4-mini

\tcbline

\textbf{System prompt:}
\par 
You are a biomedical researcher briefly explaining a strategy of how to identify novel therapeutic compounds to test using this assay to treat \{disease\_name\}.

\tcbline 

\textbf{User prompt:}
\par 
Here is a proposed experimental assay identified for treating \textquotesingle\{disease\_name\}\textquotesingle:

\quad \quad Assay Name: "\{assay\_name\}"

\quad \quad Synthesize a concise and specific research goal for the *next* stage, which is focused on **identifying novel therapeutic compounds** to test using this assay to treat \{disease\_name\}.

\quad \quad It's important that you connect the goal of this assay to how it is important for **identifying** novel therapeutic compounds to treat \{disease\_name\}.

\quad \quad Provide ONLY the synthesized goal string as the response.

\end{tcolorbox}

\textbf{Supplementary Figure \thesuppfigure}{ Prompt for synthesizing goal for candidate generation}
\label{fig:prompt_goal_candidate_generation}
\end{suppfigureenv}

\clearpage
\subsubsection{Therapeutic Candidate Generation}

\begin{suppfigureenv}
\centering
\begin{tcolorbox}[
    enhanced, 
    colback=black!5!white,   
    colframe=black!75!black, 
    arc=3mm,                
    boxrule=1pt,
]

\textbf{Model:} o4-mini

\tcbline

\textbf{System prompt:}
\par 
You are an expert drug development researcher focused on generating high-quality, specific, testable **drug candidates**.

\quad Your goal is to propose novel, single-agent drug candidates (specific molecules, potentially repurposed, that are commercially available; mention catalog numbers if possible).

\quad You are interested in finding candidates with:

\quad 1.  **Strong Target Validation:** The target pathway/mechanism has robust evidence (genetic, functional) linking it *specifically* to the disease pathophysiology.

\quad 2.  **Relevant Preclinical/Clinical Evidence:** Supporting data exists from disease-relevant models or, ideally, preliminary human data (even if from related conditions or pilot studies).

\quad 3.  **Mechanistic Specificity:** A clear, well-defined molecular mechanism is preferable over broad, non-specific actions.

\quad 4.  **Novelty (Balanced with Validation):** Innovative, exciting, and novel approaches that can advance treatment for \{disease\_name\}, that are also grounded in strong scientific rationale and evidence.

\quad - Focus on compounds that are commercially available (mention catalog numbers) and can be developed into drugs

\quad - Favor mechanisms with minimal impact on other cellular functions

\quad - Compounds should be novel for treatment of the disease and ideally address novel targets for the diseased cell type (i.e. not tested in prior studies)

\tcbline 

\textbf{User prompt:}
\par 
Return a list of \{2*num\_queries\} queries (separated by \textless{}\textgreater{}) that would be useful in doing background research for the goal of \{candidate\_generation\_goal\}.

\quad These queries will be given to a highly-trained research team to investigate the scientific literature in depth, so the queries should be able to capture broadly relevant information (30+ words) and search relevant literature across biomedical, clinical, and biochemical literature about the disease or therapeutic landscape. You don't need to propose specific drugs, but the queries should be able to capture relevant scientific information that may help with proposing drug candidates.

\quad You have \{2*num\_queries\} queries, so spread your queries out to cover as much ground as possible. Generate \{num\_queries\} queries to cover literature about the therapeutic landscape, especially those that can help \{candidate\_generation\_goal\} and generate \{num\_queries\} queries to cover literature about the biological and mechanistic aspects about \{disease\_name\} itself. In formatting, don't number the queries, just output a string with \{2*num\_queries\} queries separated by \textless{}\textgreater{}.

\quad Generate queries for the goal of \{candidate\_generation\_goal\} that actively seek and prioritize:

\quad *\ \ \ **Target Validation:** Studies demonstrating the target pathway's dysregulation or causative role.

\quad *\ \ \ **Efficacy in Relevant Models:** Evidence of the drug candidate's efficacy in cell or animal models that *closely mimic* disease pathology, or in patient-derived cells. Prefer this over general mechanism-of-action studies.

\quad *\ \ \ **Mechanism Confirmation:** Studies confirming the candidate engages the target and modulates the specific pathway *as hypothesized* in a relevant biological system.

\quad *\ \ \ **Pharmacokinetics/Safety Data:** Existing data on the candidate's ADME properties, safety profile (especially human data), and known ability to reach relevant tissues.

\end{tcolorbox}

\textbf{Supplementary Figure \thesuppfigure}{ Prompt for generating queries for literature search}
\label{fig:prompt_cand_gen_lit_search}
\end{suppfigureenv}

\newpage

\begin{suppfigureenv}
\centering
\begin{tcolorbox}[
    enhanced, 
    colback=black!5!white,   
    colframe=black!75!black, 
    arc=3mm,                
    boxrule=1pt,
]

\textbf{Model:} o4-mini

\tcbline

\textbf{System prompt:}
\par 
You are an expert drug development researcher focused on generating high-quality, specific, testable **drug candidates**. Your task is to generate novel, testable therapeutic candidates for \{disease\_name\}, based on the provided research goal and background literature.

\quad You are interested in discovering and proposing candidates with:

\quad 1.  **Strong Target Validation:** The target pathway/mechanism has robust evidence (genetic, functional) linking it *specifically* to the disease pathophysiology.

\quad 2.  **Relevant Preclinical/Clinical Evidence:** Supporting data exists from disease-relevant models or, ideally, preliminary human data (even if from related conditions or pilot studies).

\quad 3.  **Developmental Feasibility:** The candidate has known properties that facilitate development.

\quad 4.  **Mechanistic Specificity:** A clear, well-defined molecular mechanism is preferable over broad, non-specific actions.

\quad 5.  **Novelty (Balanced with Validation):** Innovative approaches are valued, but must be grounded in strong scientific rationale and evidence.

\quad For EACH hypothesis object in the \texttt{\textasciigrave hypotheses\textasciigrave} array, you MUST provide ALL of the following fields:

\quad \quad 1.  \texttt{\textasciigrave candidate\textasciigrave}: The specific drug/therapeutic proposed. Must be a single agent, not a combination. Do not propose special formulations or delivery methods.

\quad \quad 2.  \texttt{\textasciigrave hypothesis\textasciigrave}: A specific, compelling mechanistic hypothesis detailing how the candidate (a commercially available compound, mention catalog numbers if applicable) will treat \texttt{\textasciigrave\detokenize{{disease_name}}\textasciigrave} at a molecular/cellular level.

\quad \quad 3.  \texttt{\textasciigrave reasoning\textasciigrave}: Detailed scientific reasoning, explaining the mechanistic rationale, evidence, translational considerations, target validation, and novelty of the candidate.

\quad **Output Format Specification (Strict Adherence Required):**

\quad Your entire output MUST be a text block. Generate exactly \{num\_candidates\} candidate proposals.

\quad Each candidate proposal MUST start with "\textless{}CANDIDATE START\textgreater{}" on a new line and end with "\textless{}CANDIDATE END\textgreater{}" on a new line.

\quad Within each block, each piece of information (CANDIDATE, HYPOTHESIS, REASONING.) MUST start on a new line, beginning with its EXACT header (e.g., \texttt{\textasciigrave CANDIDATE:\textasciigrave}, \texttt{\textasciigrave HYPOTHESIS:\textasciigrave}) followed by the content.

\quad Do NOT include any other text before the first "\textless{}CANDIDATE START\textgreater{}" or after the last "\textless{}CANDIDATE END\textgreater{}".

\quad Example for one candidate (repeat this block structure \{num\_candidates\} times):\par\noindent

\quad \textless{}CANDIDATE START\textgreater{}\par\noindent

\quad CANDIDATE: The specific drug/therapeutic proposed. Must be a single agent, not a combination. Do not propose special formulations or delivery methods.\par\noindent

\quad HYPOTHESIS: A specific, compelling mechanistic hypothesis detailing how the candidate (a commercially available compound, mention catalog numbers if applicable) will treat \{disease\_name\} at a molecular/cellular level.\par\noindent

\quad REASONING: Detailed scientific reasoning, explaining the mechanistic rationale, evidence, translational considerations, target validation, and novelty of the candidate.\par\noindent

\quad \textless{}CANDIDATE END\textgreater{}

\tcbline 

\textbf{User prompt:}
\par 
Generate exactly **\{num\_candidates\}** distinct and scientifically rigorous proposals for therapeutic candidates to treat \{disease\_name\}. Here is some relevant background information that can guide your proposals:
\quad \{therapeutic\_candidate\_review\_output\}

\end{tcolorbox}

\textbf{Supplementary Figure \thesuppfigure}{ Prompt for generating therapeutic candidate hypotheses}
\label{fig:prompt_gen_thera_cand_hypo}
\end{suppfigureenv}

\newpage

\newpage

\begin{suppfigureenv}
\centering
\begin{tcolorbox}[
    enhanced, 
    colback=black!5!white,   
    colframe=black!75!black, 
    arc=3mm,                
    boxrule=1pt,
]

\textbf{Model:} Falcon

\tcbline

You are an expert drug development team leader focused on validating **drug candidates**. Your task is to evaluate promising repurposed drugs or therapeutic candidates for \{disease\_name\} proposed by your research team.

\quad \quad Given the following therapeutic candidate, do a comprehensive literature review to evaluate if this therapeutic candidate has potential for \{disease\_name\}. Search relevant literature across

\quad \quad biomedical, clinical, and biochemical literature about the disease or proposed therapeutic. This must be very detailed and comprehensive, as it will determine the direction of the team.

\{Candidate\}

Provide your response in the following format, like an evaluation for a scientific proposal:

\quad Overview of Therapeutic Candidate: Explain the natural or synthetic origins of this therapeutic candidate, including how it was synthesized or discovered. Explain which class of therapeutic compounds this belongs to, and how this class of compounds has previously been used in general.

\quad Therapeutic History: Summarize previous biochemical, clinical or veterinary uses of this drug or drug class, if any. Examine to see if the therapeutic candidate has ever been used for treating \{disease\_name\} or any similar disease.

\quad Mechanism of Action: Explain the known mechanism(s) of action of this compound to the full extent of molecular detail that is known. Explain the biochemistry and molecular interactions of the therapeutic candidate in any relevant literature.

\quad Expected Effect: Explain the expected effect of this compound in the assay proposed and the mechanism by which it will act. If the drug is acting on proteins, reference literature which shows this gene is expressed in the cell type being assayed.

\quad Overall Evaluation: Give your overall thoughts on this therapeutic candidate. Include strengths and weaknesses of this therapeutic candidate for treating \{disease\_name\}.

\end{tcolorbox}
\textbf{Supplementary Figure \thesuppfigure}{ Prompt for generating detailed investigation for each therapeutic candidate}
\label{fig:prompt_detailed_investigation_thera_cand}
\end{suppfigureenv}

\begin{suppfigureenv}
\centering
\begin{tcolorbox}[
    enhanced, 
    colback=black!5!white,   
    colframe=black!75!black, 
    arc=3mm,                
    boxrule=1pt,
]

\textbf{Model:} Finch

\tcbline

Analyze the zipped .fcs files from three 96-well plates to measure average Alexa Fluor 647 signal in live retinal pigment epithelial cells across different drug treatments.

Use the metadata file to match well positions (e.g., A1) with their corresponding drug treatments and plate information. Use fcsparser to parse the files. 

Create appropriate gates using the "DMSO control + beads + cells + DAPI" well to isolate the live, singlet cell population. Make the necessary plots to identify the required gating thresholds. I recommend using polygon coordinates to remove debris, dead cells, and doublets or aggregates. Make sure to exclude all debris, and that the singlet gating is not too strict. Once you have the best gating strategy possible that gives reasonable percentage gated events, check the gating on the other two control wells to make sure comparable percentage gated events. 

Determine if there are plate-to-plate effect, if there is, take appropriate normalization steps using the DMSO control before downstream analysis. 

Measure the Alexa Fluor 647 signal intensity in these cells, and compile the results into a CSV file. 

Determine if there are any drug treatments that significantly increased signal compared to control. I recommend using the Dunnett test to compare the drug treatments to the DMSO control. Generate a barplot showing the average Alexa Fluor 647 fluorescence values for each drug treatment with error bars showing the standard error of the mean, include highlighting any drugs that significantly increase Alexa Fluor 647 signal using asterisks to denote significance. Exclude any control treatments in the final plot. 

Do not stop early. Make sure you have processed all files available.

\end{tcolorbox}
\textbf{Supplementary Figure \thesuppfigure}{ Prompt for performing flow cytometry analyses from zipped .fcs files}
\label{fig:finch_prompt}
\end{suppfigureenv}

\clearpage

\begin{tcolorbox}[
    enhanced, 
    breakable,
    colback=black!5!white,   
    colframe=black!75!black, 
    arc=3mm,                
    boxrule=1pt,
]

\textbf{Model:} Claude 3.7 Sonnet

\tcbline
\textbf{System prompt:}
\par 
You are an experienced drug development committee member, with broad scientific expertise.

\quad Your objective is to rigorously compare two preclinical drug candidate proposals. Your primary goal is to select the hypothesis with the **highest probability of successful experimental outcome AND eventual translation into a viable therapy** for \{disease\_name\}.

\quad Your evaluation must be based strictly on the presented scientific evidence, scientific novelty, methodological rigor, and logical interpretation. Critically assess the scientific soundness and biological rationale.

\quad **Prioritize your evaluation based on these key criteria, reflecting human expert preferences:**

\quad 1.  **Strength and Relevance of Supporting Evidence (Highest Priority):**

\quad \quad *\ \ \ **Existing Data:** Is there robust existing data (preclinical, clinical, especially for \{disease\_name\} or highly relevant biological contexts) supporting the hypothesis? Give significant weight to positive data from later-stage research (e.g., in vivo, clinical trials for the drug or drug class, especially if successful for related conditions).

\quad \quad *\ \ \ **Relevance to the Problem:** Is the evidence *directly relevant* to the proposed mechanism in the context of \{disease\_name\} and the specific biological problem being addressed? General efficacy in unrelated conditions is less compelling.

\quad \quad *\ \ \ **Negative Evidence:** Are there known failures, lack of efficacy, or contradictory data for this drug/class in relevant contexts? This is a strong deterrent.

\quad \quad *\ \ \ **Quality of References:** Are the cited references (if any are explicitly mentioned or implied by the reasoning) strong and relevant to the drug and its proposed action in this context?

\quad 2.  **Mechanism of Action (MoA) - Clarity, Plausibility, and Specificity for \{disease\_name\}:**

\quad \quad *\ \ \ **Clarity \&{} Detail:** Is the MoA clearly articulated, scientifically sound, detailed, and plausible? Vague mechanisms are less favored.

\quad \quad *\ \ \ **Directness:** How *direct* is the MoA to addressing the core pathology of \{disease\_name\}? Prefer hypotheses targeting core, upstream issues relevant to the disease's cellular or molecular basis over indirect or downstream effects, unless the indirect effect is exceptionally well-supported and highly plausible.

\quad \quad *\ \ \ **Specificity \&{} Target Biology:** Is the therapeutic target well-defined? Is its biology central to the pathogenesis of \{disease\_name\}? Specific pathway/target engagement is preferred over overly broad actions that might increase risk.

\quad 3.  **Safety, Tolerability, and Risk Profile (High Priority):**

\quad \quad *\ \ \ **Known Safety Profile:** What is the known safety profile of the drug or drug class? Candidates with established clinical safety (e.g., repurposed drugs with a *strong rationale* for \{disease\_name\}) are highly favored.

\quad \quad *\ \ \ **Off-Target Effects/Toxicity:** Are there significant concerns about off-target effects, general toxicity, or specific organ/system toxicity relevant to \{disease\_name\} or its treatment? This is a major red flag. Assess potential for on-target and off-target liabilities.

\quad \quad *\ \ \ **Pleiotropic Effects:** Consider if broad (pleiotropic) effects are beneficial or detrimental in this specific disease context.

\quad 4.  **Feasibility of Experimental Plan and Drug Delivery:**

\quad \quad *\ \ \ **Methodological Rigor:** Does the proposal (even if brief) outline a clear, methodologically sound, and feasible experimental plan? Are specific, measurable outcomes related to the therapeutic goal for \{disease\_name\} defined?

\quad \quad *\ \ \ **Model System Relevance:** Is the proposed model system (e.g., relevant cell lines, animal models of \{disease\_name\}) appropriate and predictive?

\quad \quad *\ \ \ **Drug Delivery to Target:** Crucially, assess the feasibility of drug delivery to the target tissue or system. Consider physicochemical properties and ADME/PK profiles (bioavailability, metabolic stability) in terms of achieving therapeutic concentrations with an acceptable therapeutic window. Is the proposed route of administration plausible and supported for this drug type and disease?

\quad 5.  **Scientific Novelty (Balanced with Evidence and Safety):**

\quad \quad *\ \ \ Does the hypothesis offer a novel scientific advancement for \{disease\_name\}? Novelty is valued, but *only if supported by plausible science and a reasonable safety outlook*. Do not prefer novelty if it comes at the cost of strong evidence for a more established, safer approach. An innovative approach to an underexplored but plausible pathway can be positive.

\quad **Synthesize these factors:**

\quad While detailed pharmacological properties (in vitro potency, selectivity, SAR, chemical liabilities) and ADME/PK parameters (CYP inhibition/induction, DDI risk) are important, **focus on their *implications* for efficacy, safety, and translatability, as a human expert panel would.** Avoid getting lost in excessive technical detail unless it directly and significantly impacts one of the prioritized criteria (e.g., a PK issue preventing target engagement would be critical).

\quad The goal of this task is to choose the best therapeutic candidate that has the most potential for treating \{disease\_name\}.

\tcbline 

\textbf{User prompt:}
\par 
Evaluate the preclinical drug candidate data package using the structure below. This evaluation informs a critical decision.

\quad **Respond ONLY in the specified JSON format, do not include any text outside the JSON object itself.**

\begin{Verbatim}[breaklines]
{{
"Analysis": {"type": "string", "description": "[Provide a detailed analysis of the two drug candidates, based on the evaluation criteria and the evidence provided.]"},
"Reasoning": {"type": "string", "description": "[Choose which drug candidate is better. Provide a detailed explanation for why you think the winner is better than the loser, based on the evaluation criteria and the evidence provided.]"},
"Winner": {"type": "string", "description": "[Return the name and ID number of the candidate that you think is better between the two candidates, as a tuple. It should be formatted as (winner_name, winner_id)]"},
"Loser": {"type": "string", "description": "[Return the name and ID number of the candidate that you think is worse between the two candidates, as a tuple. It should be formatted as (loser_name, loser_id)]"},
}}
\end{Verbatim}

\end{tcolorbox}
\centering{
\textbf{Supplementary Figure \thesuppfigure} {Prompt for LLM judge to rank therapeutic candidate hypotheses}
}

\clearpage

\begin{suppfigureenv}
\centering
\includegraphics[width=0.8\columnwidth]{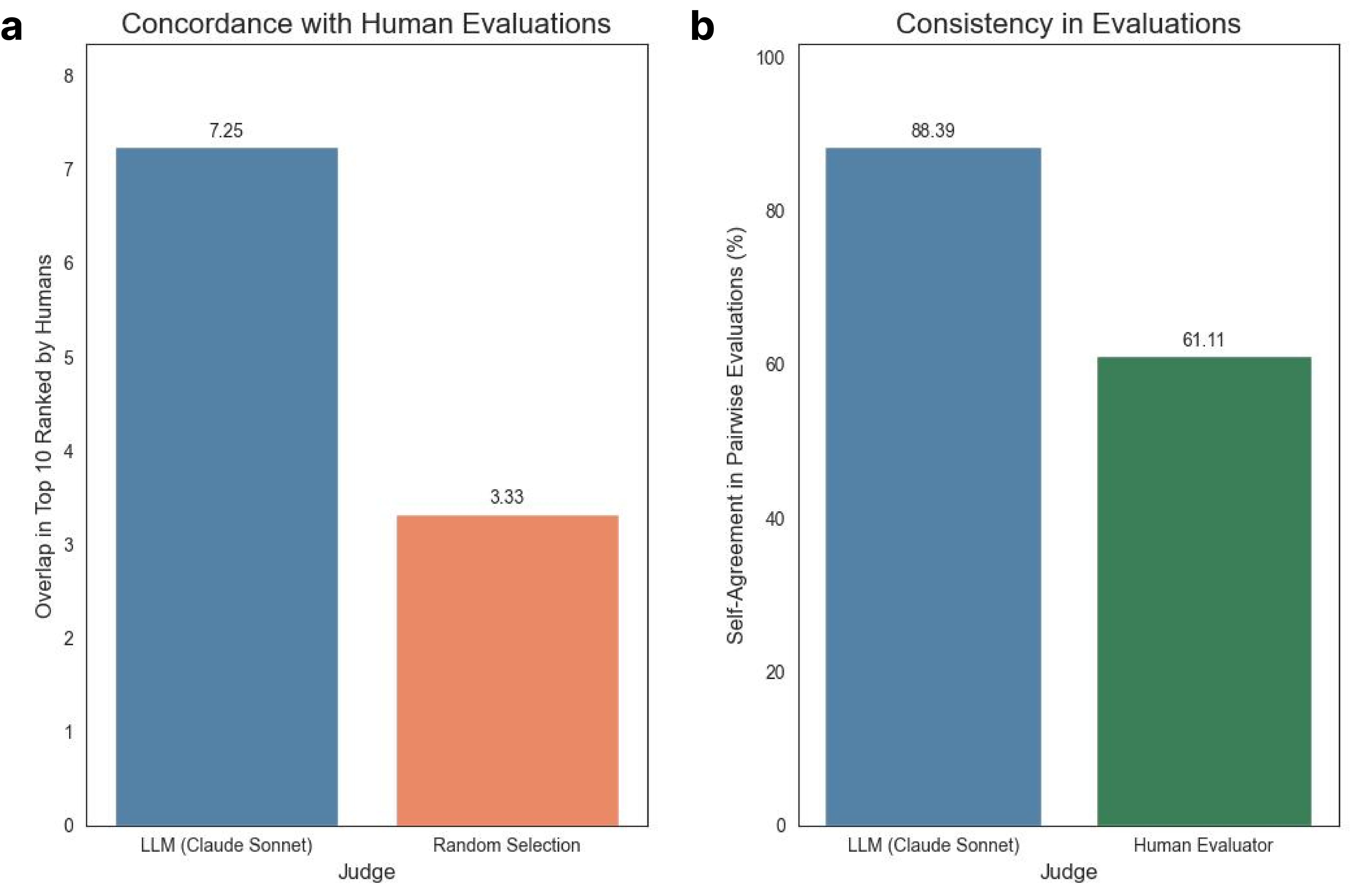}

\textbf{Supplementary Figure \thesuppfigure} {Robin LLM judge preferences align with human expert evaluations. A) The Robin LLM judge preferences align well with human expert evaluations, selecting an average of 7.25 of their top 10 hypotheses. B) The Robin LLM judge is more consistent in its preference when selecting between two hypotheses than human experts.}
\label{fig:humanevals}
\end{suppfigureenv}

\begin{suppfigureenv}
\centering
\includegraphics[width=0.6\columnwidth]{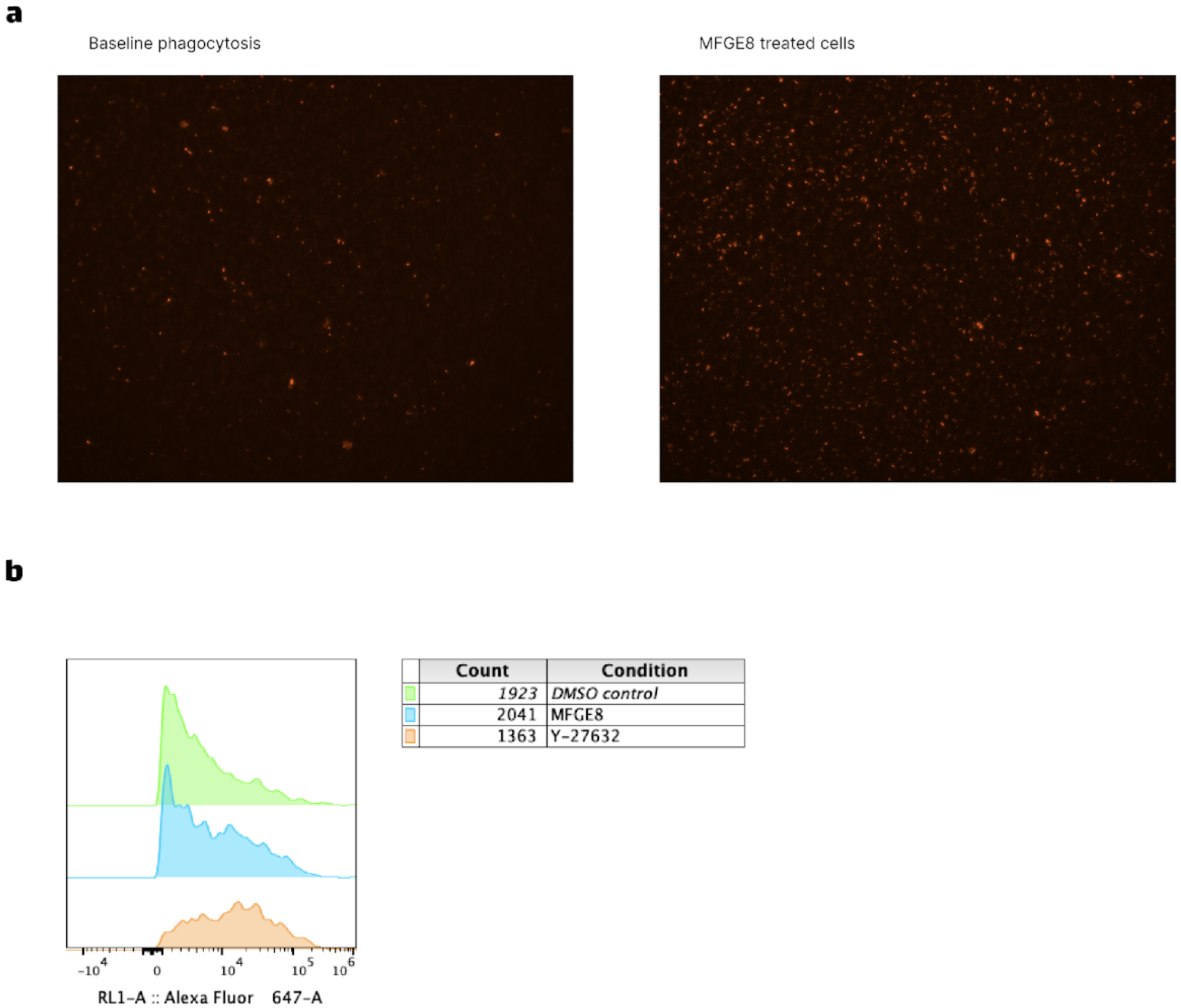}

\textbf{Supplementary Figure \thesuppfigure}{ (A) Fluorescence microscopy images of ARPE-19 cells treated with DMSO vehicle (left) or MFGE8 (positive control; right). (B) Flow cytometry histograms showing the fluorescence intensity of ARPE-19 cells treated with the DMSO vehicle, MFGE8 or the ROCK inhibitor Y-27632.}
\label{fig:sup_microscopy_bead_uptake}
\end{suppfigureenv}

\newpage

\begin{suppfigureenv}
\centering
\includegraphics[width=0.8\columnwidth]{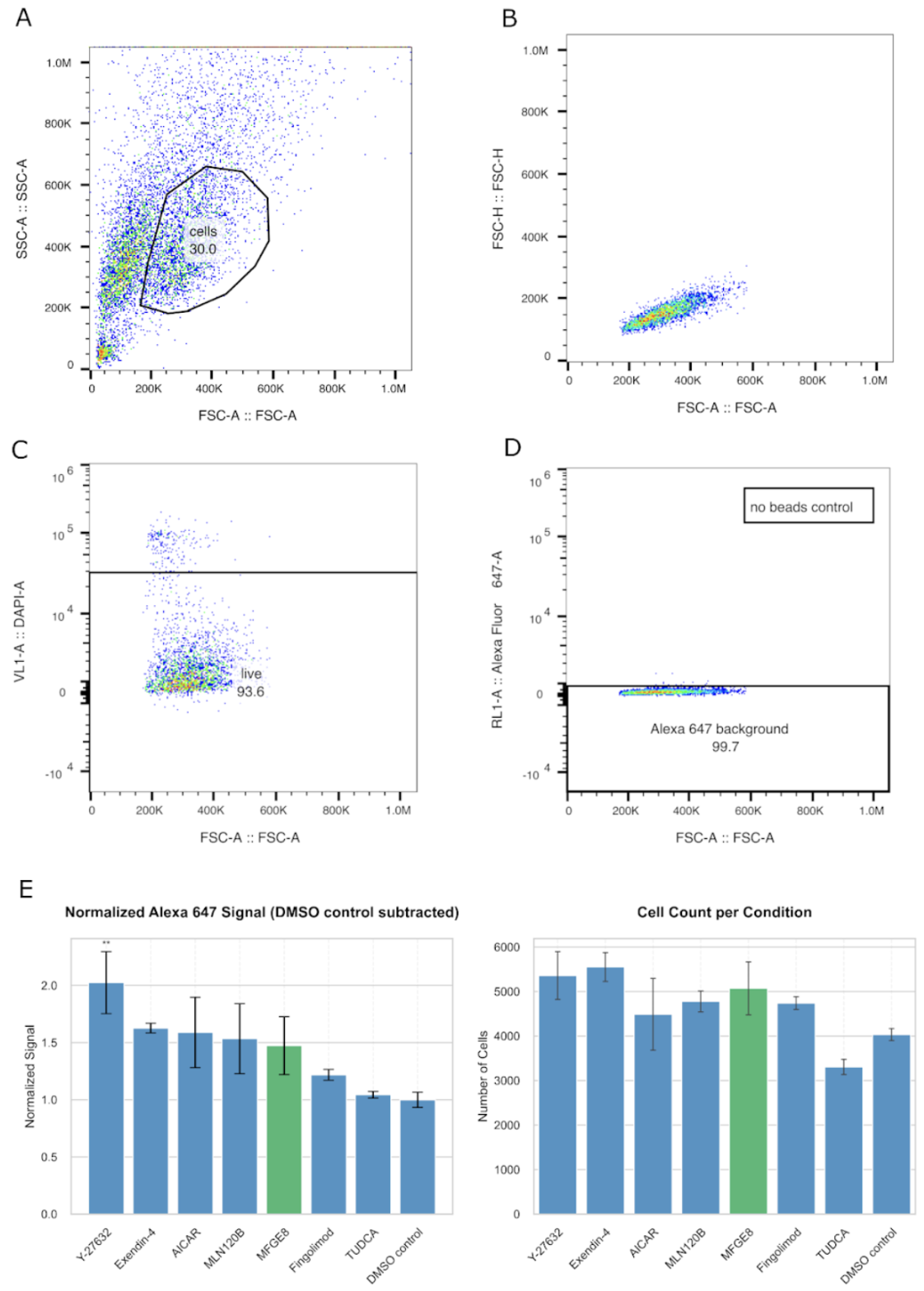}

\textbf{Supplementary Figure \thesuppfigure}{ Human analysis of round 1 flow cytometry data. (A-D) gating strategy. (A) The main cell population is identified. (B) Singlets are identified. (C) Dead cells are gated out (D) Alexa 647 background is gated out. (E) Barplots showing normalized gross alexa 647 signal, net alexa 647 signal (DMSO control subtracted) and cell count per condition (after passing all gates). Error bars show the standard error of the mean. Asterisks denote statistical significance: p$<$0.05 (*), p$<$0.01 (**), p$<$0.001 (***) (Dunnett’s test).}
\label{fig:sup_flow_cytometry_round1}
\end{suppfigureenv}

\newpage

\begin{suppfigureenv}
\centering
\includegraphics[width=0.8\columnwidth]{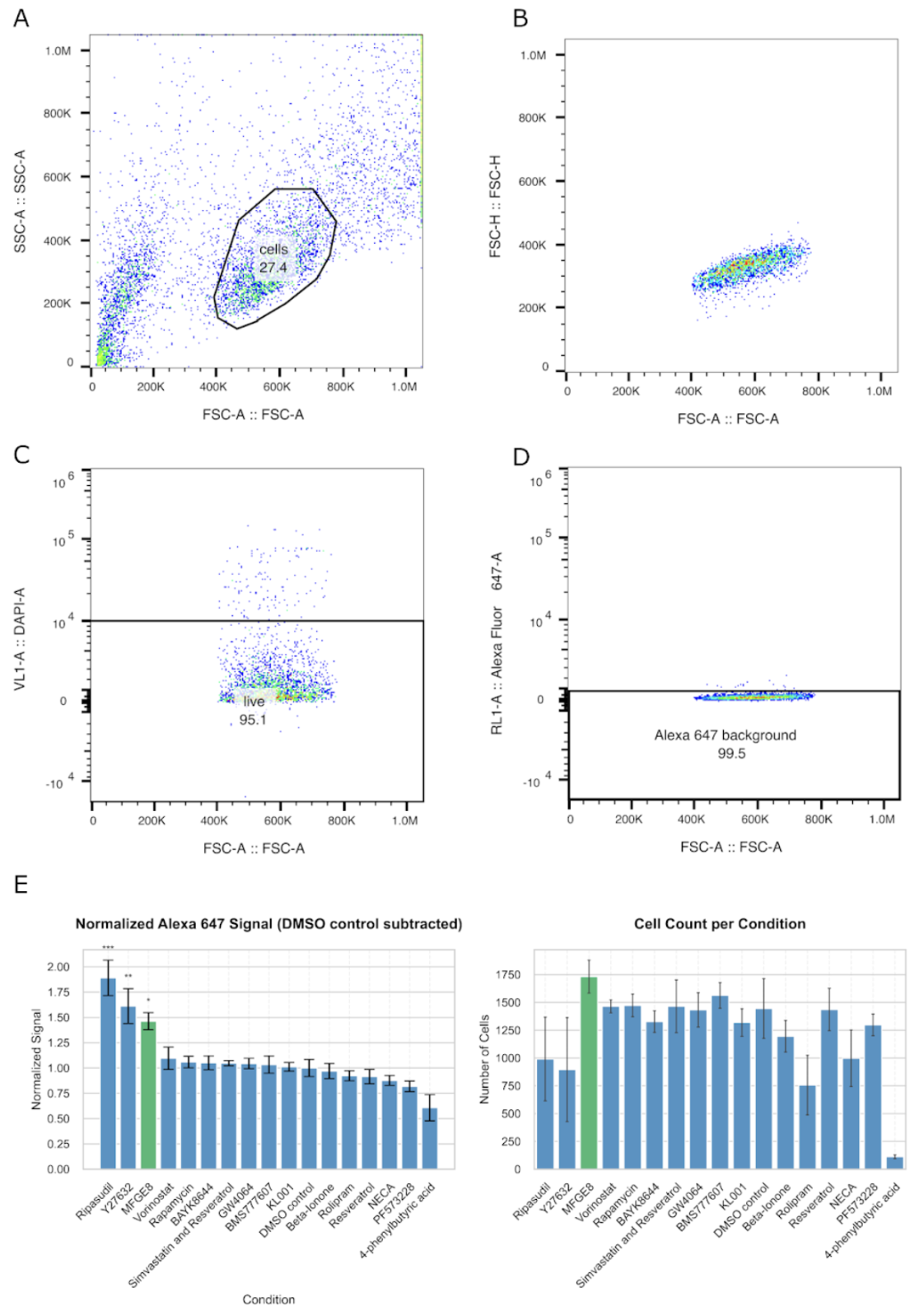}

\textbf{Supplementary Figure \thesuppfigure}{ Human analysis of round 2 flow cytometry data. (A-D) gating strategy. (A) The main cell population is identified. (B) Singlets are identified. (C) Dead cells are gated out (D) Alexa 647 background is gated out. (E) Barplots showing normalized gross alexa 647 signal, net alexa 647 signal (relative to DMSO control) and cell count per condition (after passing all gates). Error bars show the standard error of the mean. Asterisks denote statistical significance: p$<$0.05 (*), p$<$0.01 (**), p$<$0.001 (***) (Dunnett’s test).}
\label{fig:sup_flow_cytometry_round2}
\end{suppfigureenv}

\begin{suppfigureenv}
\centering
\includegraphics[width=\columnwidth]{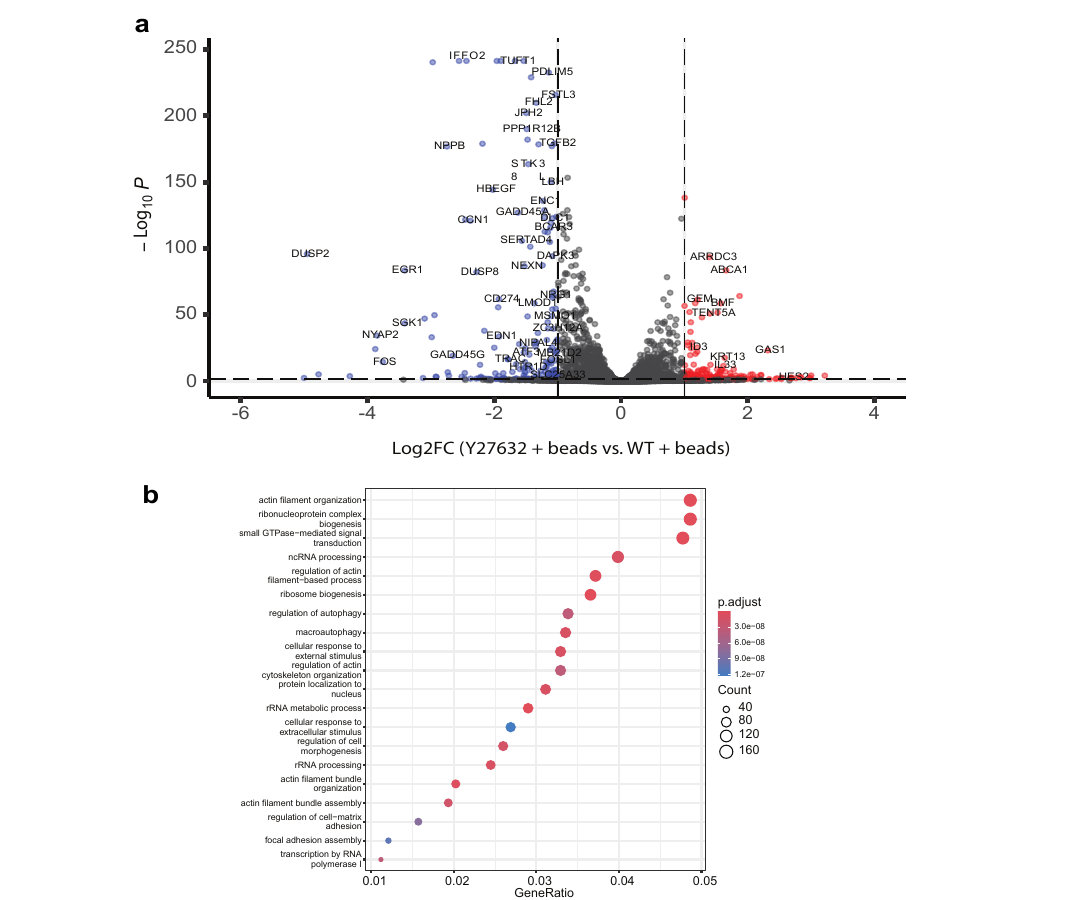}

\textbf{Supplementary Figure \thesuppfigure}{Human analysis of the RNA sequencing experiment (A) Volcano plot showing differential gene expression between ARPE-19 cells treated with Y-27632 versus vehicle control (DMSO). The x-axis represents log$_{2}$ fold change in expression, with positive values indicating up-regulation and negative values indicating down-regulation. The y-axis shows –log$_{10}$ of the p-value for each gene, highlighting statistical significance. Genes that pass the significance threshold (|log$_{2}$FC| > 1, p < 0.05) are colored (red for up-regulated, blue for down-regulated), while non-significant genes are displayed in gray. (B) Gene Ontology (GO) enrichment analysis (using biological process terms) of differentially expressed genes performed using the ReactomePA package. The y-axis lists significantly enriched GO terms, while the x-axis shows the gene ratio, calculated as the number of input genes associated with each GO term divided by the total number of input genes. Dot size corresponds to the number of genes mapped to each term, and color indicates the adjusted p-value, reflecting statistical significance of the enrichment.
}
\label{fig:human_RNAseq}
\end{suppfigureenv}

\newpage

\noindent
\begin{minipage}[t]{0.48\textwidth}
\begin{tcolorbox}[colback=black!5!white, colframe=black!65!white, title=Polycystic Ovary Syndrome]
\raggedright
\begin{enumerate}
  \item N-Acetylcysteine
  \item Sitagliptin
  \item Empagliflozin
  \item INT131
  \item Dihydroberberine
  \item MSDC-0602K
  \item Dapagliflozin
  \item BGP-15
  \item Trodusquemine
  \item Anakinra
\end{enumerate}
\end{tcolorbox}
\vspace{1mm}
\suppcaptionbox{fig:pcos}{10 Therapeutic Candidates Proposed by Robin for Polycystic Ovary Syndrome}
\end{minipage}%
\hfill
\begin{minipage}[t]{0.48\textwidth}
\begin{tcolorbox}[colback=black!5!white, colframe=black!65!white, title=Celiac Disease]
\raggedright
\begin{enumerate}
  \item ZED1227
  \item BL-7010
  \item Latiglutenase
  \item AMG 714
  \item Tofacitinib
  \item Aspergillus niger prolyl endopeptidase
  \item Bifidobacterium longum CECT 7347
  \item 4-Phenylbutyric acid
  \item Lactobacillus casei ATCC 9595
  \item Epigallocatechin gallate
\end{enumerate}
\end{tcolorbox}
\vspace{1mm}
\suppcaptionbox{fig:celiac}{10 Therapeutic Candidates Proposed by Robin for Celiac Disease}
\end{minipage}

\noindent
\begin{minipage}[t]{0.48\textwidth}
\begin{tcolorbox}[colback=black!5!white, colframe=black!65!white, title=Charcot-Marie-Tooth Disease]
\raggedright
\begin{enumerate}
  \item CKD-504
  \item Recombinant Human Neuregulin-1 $\beta$-1 EGF-like Domain
  \item TUDCA
  \item Fasudil
  \item Guanabenz
  \item Arimoclomol
  \item Ricolinostat
  \item 4-Aminopyridine
  \item GSK269962A
  \item Y-27632
\end{enumerate}
\end{tcolorbox}
\vspace{1mm}
\suppcaptionbox{fig:cmt}{10 Therapeutic Candidates Proposed by Robin for Charcot-Marie-Tooth Disease}
\end{minipage}%
\hfill
\begin{minipage}[t]{0.48\textwidth}
\begin{tcolorbox}[colback=black!5!white, colframe=black!65!white, title=Idiopathic Pulmonary Fibrosis]
\raggedright
\begin{enumerate}
  \item Pamrevlumab
  \item GSK3008348
  \item BI 1015550
  \item GB0139
  \item KD025
  \item GKT137831
  \item CC-90001
  \item BMS-986278
  \item X203
  \item Dimethyl Fumarate
\end{enumerate}
\end{tcolorbox}
\vspace{1mm}
\suppcaptionbox{fig:ipf}{10 Therapeutic Candidates Proposed by Robin for Idiopathic Pulmonary Fibrosis}
\end{minipage}

\noindent
\begin{minipage}[t]{0.48\textwidth}
\begin{tcolorbox}[colback=black!5!white, colframe=black!65!white, title=Non-alcoholic Steatohepatitis]
\raggedright
\begin{enumerate}
  \item Resmetirom
  \item Aldafermin
  \item Saroglitazar
  \item Pegbelfermin
  \item VK2809
  \item Tropifexor
  \item Cilofexor
  \item EDP-305
  \item Pemafibrate
  \item PXL770
\end{enumerate}
\end{tcolorbox}
\vspace{1mm}
\suppcaptionbox{fig:nash}{10 Therapeutic Candidates Proposed by Robin for Non-alcoholic Steatohepatitis}
\end{minipage}%
\hfill
\begin{minipage}[t]{0.48\textwidth}
\begin{tcolorbox}[colback=black!5!white, colframe=black!65!white, title=Sarcopenia]
\raggedright
\begin{enumerate}
  \item Bimagrumab
  \item Tirasemtiv
  \item Reldesemtiv
  \item SS-31
  \item S48168
  \item Urolithin A
  \item Setanaxib
  \item Elamipretide
  \item JTV-519
  \item Rolipram
\end{enumerate}
\end{tcolorbox}
\vspace{1mm}
\suppcaptionbox{fig:sarcopenia}{10 Therapeutic Candidates Proposed by Robin for Sarcopenia}
\end{minipage}

\noindent
\begin{minipage}[t]{0.48\textwidth}
\begin{tcolorbox}[colback=black!5!white, colframe=black!65!white, title=Age-related hearing loss]
\raggedright
\begin{enumerate}
  \item Nicotinamide Riboside
  \item N-acetylcysteine
  \item Tauroursodeoxycholic Acid
  \item Metformin
  \item Rapamycin
  \item MitoQ
  \item Etanercept
  \item Resveratrol
  \item SKQ1
  \item Quercetin
\end{enumerate}
\end{tcolorbox}
\vspace{1mm}
\suppcaptionbox{fig:arhl}{10 Therapeutic Candidates Proposed by Robin for Age-related hearing loss}
\end{minipage}%
\hfill
\begin{minipage}[t]{0.48\textwidth}
\begin{tcolorbox}[colback=black!5!white, colframe=black!65!white, title=Glaucoma]
\raggedright
\begin{enumerate}
  \item Netarsudil
  \item Y-27632
  \item Ripasudil
  \item ISTH0036
  \item Fasudil
  \item Rapamycin
  \item Verteporfin
  \item Pirfenidone
  \item KD025
  \item Losartan
\end{enumerate}
\end{tcolorbox}
\vspace{1mm}
\suppcaptionbox{fig:glaucoma}{10 Therapeutic Candidates Proposed by Robin for Glaucoma}
\end{minipage}

\noindent
\begin{minipage}[t]{0.48\textwidth}
\begin{tcolorbox}[colback=black!5!white, colframe=black!65!white, title=Friedreich's Ataxia]
\raggedright
\begin{enumerate}
  \item Deferiprone
  \item Omaveloxolone (RTA 408)
  \item EPI-743
  \item Leriglitazone
  \item Elamipretide
  \item SS-31
  \item RG2833
  \item Dimethyl fumarate
  \item KH176
  \item Nicotinamide riboside
\end{enumerate}
\end{tcolorbox}
\vspace{1mm}
\suppcaptionbox{fig:fa}{10 Therapeutic Candidates Proposed by Robin for Friedreich's Ataxia}
\end{minipage}%
\hfill
\begin{minipage}[t]{0.48\textwidth}
\begin{tcolorbox}[colback=black!5!white, colframe=black!65!white, title=Chronic Kidney Disease]
\raggedright
\begin{enumerate}
  \item OLT1177
  \item VX-765
  \item Tranilast
  \item MCC950
  \item Brilliant Blue G
  \item Disulfiram
  \item CY-09
  \item A-438079
  \item Oridonin
  \item INF4E
\end{enumerate}
\end{tcolorbox}
\vspace{1mm}
\suppcaptionbox{fig:ckd}{10 Therapeutic Candidates Proposed by Robin for Chronic Kidney Disease}
\end{minipage}

\end{document}